\documentclass{article}

\PassOptionsToPackage{numbers, compress}{natbib}

 \usepackage[main, final]{neurips_2026}

\usepackage[utf8]{inputenc} %
\usepackage[T1]{fontenc}    %
\usepackage{hyperref}       %
\usepackage{url}            %
\usepackage{booktabs}       %
\usepackage{amsfonts}       %
\usepackage{nicefrac}       %
\usepackage[table]{xcolor}  %
\usepackage{graphicx}
\usepackage{amsmath}
\usepackage{float}
\usepackage{lipsum}
\usepackage{multirow}
\usepackage{enumitem}
\usepackage{amssymb}
\usepackage{algorithm}
\usepackage{algpseudocode}
\usepackage{wrapfig}
\usepackage{array}
\usepackage{subcaption}
\usepackage{caption}
\usepackage{ragged2e} %
\usepackage{tabularx}
\usepackage{colortbl}
\usepackage[table]{xcolor}
\usepackage{makecell}
\usepackage{placeins}
\usepackage[normalem]{ulem} %
\usepackage{bm}
\useunder{\uline}{\ul}{}

\definecolor{groupgray}{gray}{0.93}

\title{Seeing the Unseen: Unified Visible–Invisible Motion for Physically Consistent Video Generation
}

\author{%
Mingyang Bao$^1$, Fangda Ye$^1$, Xihao Chen$^1$, Bobo Li$^1$, Jason Li$^2$, Shengqiong Wu$^3$\thanks{Corresponding author.} \\
  $^1$National University of Singapore, $^2$Nanyang Technological University, $^3$University of Oxford\\
  \texttt{Project page: \url{https://vid-gen.github.io/seeing_the_unseen}} \\
}

\begin{document}

\maketitle
\vspace{-0.3em}
\begin{figure}[h!]
    \centering
    \includegraphics[width=0.99\linewidth]{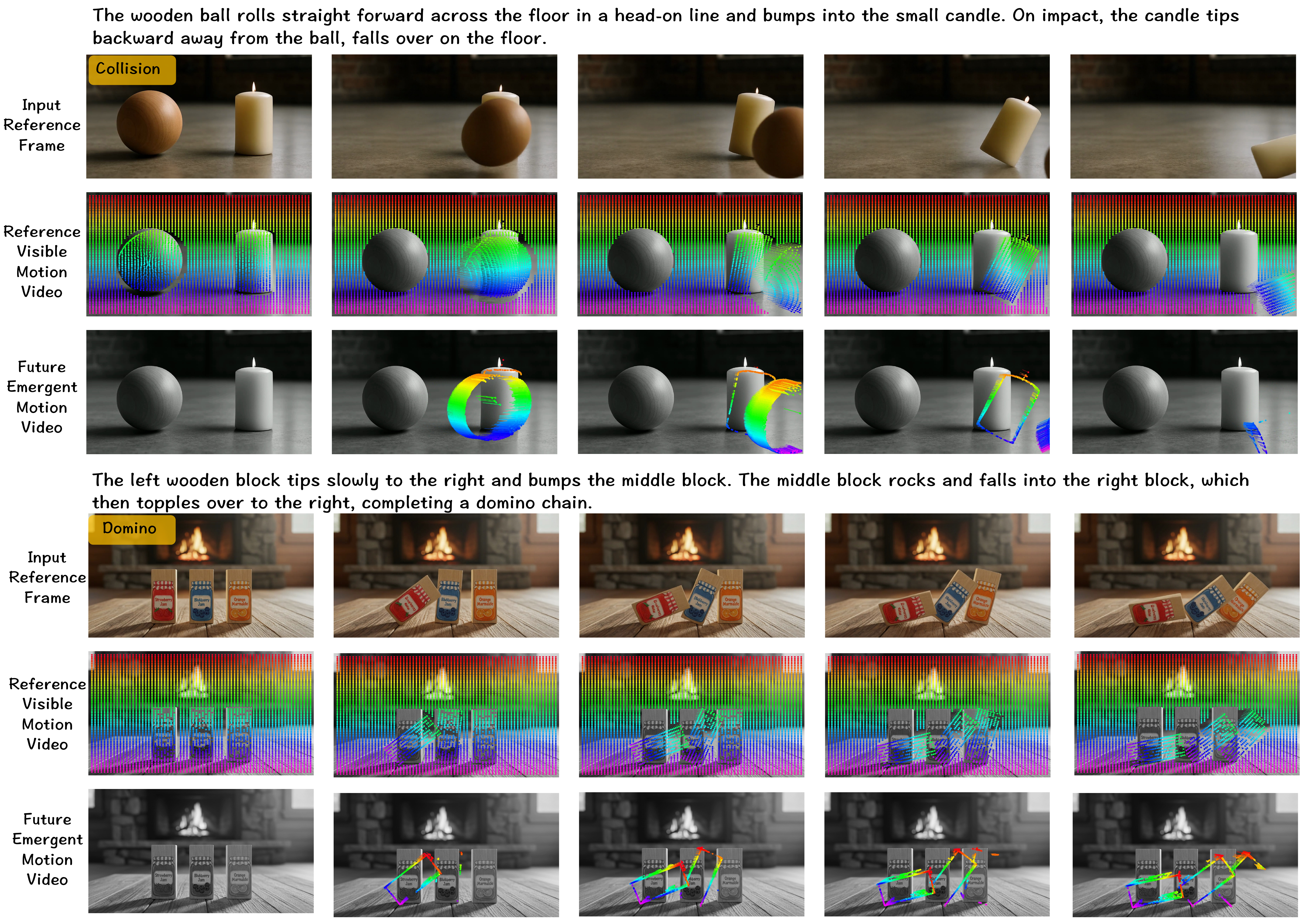}
    \caption{\textbf{Seeing the Unseen.} Our framework predicts physically consistent motion from a single image by modeling both reference-visible and future-emergent trajectories. 
By reasoning about occluded dynamics and structured interactions, it enables accurate simulation of complex events such as collisions and domino chains.}
    \label{fig:teaser}
\end{figure}

\begin{abstract}
Recent advances in video generation have achieved impressive visual quality, yet they often fail to produce physically consistent dynamics due to the lack of explicit modeling of underlying physical processes. Existing approaches either rely on simulators with strong assumptions or use coarse semantic guidance, and further struggle to effectively inject physical priors into generation models.
In this work, we propose \textbf{\textit{Seeing the Unseen}}, a unified framework for physics-grounded video generation. Our approach decomposes the problem into two stages: learning physical dynamics and injecting them into video synthesis. First, we introduce a lightweight \textit{Particle-based Graph Dynamics Simulator (PGDS)} that learns generalizable physical interactions from data and predicts plausible 3D motion trajectories. Second, we propose a \textit{Visible–Invisible Motion Field (VIMF)} that captures both motion observable in the input frame and motion that emerges over time due to object dynamics. This representation enables more complete and structured motion guidance compared to conventional physical signals.
By integrating these trajectories into a diffusion-based generator, our method produces videos that are both visually coherent and physically consistent. Extensive experiments demonstrate improved motion accuracy, interaction consistency, and reduced physical artifacts compared to strong baselines.

\end{abstract}

\section{Introduction}

Video is a fundamental medium for recording dynamic processes in the real world.
With the significant advancement of diffusion transformers~\cite{peebles2023scalable}, video generation~\citep{peebles2023scalable,yang2024cogvideox,wu2025hunyuanvideo,wan2025wan} has achieved remarkable progress in visual fidelity, temporal consistency, and complex scene synthesis. 
Recent production-level systems (e.g., Kling~\citep{kuaishou2026kling}, Gen-3~\citep{runway2024gen3}, Pika~\citep{pika2026pika2}, and Veo3.1~\cite{google2026veo31}) are capable of generating highly realistic dynamic videos and perform impressively in many everyday scenarios. 
However, the physical world is governed not only by visual appearance but also by underlying dynamics and interactions between objects, which follow well-defined physical laws. 
In contrast, existing video generation models predominantly learn pixel-level temporal statistics and lack explicit modeling of physical dynamics~\citep{bansal2025videophy,wang2025physctrl}. 
As a result, they often produce physically implausible outcomes in scenarios involving complex interactions or causal relationships, such as object interpenetration, unrealistic motion, or unstable interactions~\citep{guo2025t2vphysbench,chen2025physical,meng2024towards,bansal2024videophy}.
Improving physical consistency, therefore, requires the incorporation of effective physical priors~\citep{ji2025physmaster,liu2024physgen,huang2025vchain}. 
Yet, current approaches remain limited in both the learning of such priors and their integration into video generation models.

One line of work~\citep{liu2024physgen,chen2025physgen3d,tan2024physmotion,xie2025physanimator} learns physical priors through physics simulators~\citep{tung2023physion++,Genesis}, which explicitly model the accurate physical constraints of rigid bodies, deformable objects, or fluids under controlled settings.  
However, they typically require precise 3D reconstruction, material estimation, and physical parameter specification, which are difficult to obtain in open-world visual scenarios. 
This strong dependence on accurate scene states and simulation parameters limits their scalability and generalization.
To alleviate these requirements, some works~\citep{yuan2025newtongen,wang2025physctrl} adopt data-driven approaches to approximate physical dynamics, for instance, ordinary differential equations~\citep{yuan2025newtongen}. 
However, such methods rely on compact state representations (e.g., low-dimensional vectors), which limits their ability to capture complex interactions and multi-object dynamics.
Another line of work exploits~\citep{zhang2025think,yu2025lina,huang2025vchain,yang2025vlipp,bai2025moregen} the physical commonsense knowledge embedded in large vision-language models (LVLMs)~\cite{wang2024qwen2,openai2024gpt4o} to guide video generation at a semantic level. 
While such models can provide high-level physical reasoning, their guidance remains coarse-grained and language-centric. 
They do not directly characterize the fine-grained dynamics underlying object motion and interaction, making it difficult to impose stable and precise physical constraints during generation.

Another critical challenge is how to effectively incorporate them into video generation models. Existing methods often convert predicted motion into intermediate two-dimensional representations, such as optical flow~\citep{yuan2025newtongen,wang2025physctrl,liu2026realwonder,xie2025physanimator}, tracking videos~\citep{huang2025vchain,tan2024physmotion}, layouts~\citep{yang2025vlipp}, or sparse keypoint trajectories~\citep{gu2025diffusion,chen2025physgen3d}, and use them as conditional signals.
Although these representations can convey certain motion cues, they suffer from several limitations. 
First, projecting 3D trajectories into 2D inevitably discards depth and spatial structure, leading to compressed and incomplete physical information. 
Second, 2D motion conditions are usually restricted to regions visible from the reference view, and thus cannot describe future content that is initially occluded or unseen. 
Third, optical flow, tracking videos, and sparse keypoints only provide partial approximations of object motion, failing to preserve object-level and structure-level dynamics. 
These limitations weaken their ability to provide comprehensive physical guidance for video generation.

In this paper, we propose \textbf{Seeing the Unseen}, a unified framework for enhancing physically consistent video generation from two complementary perspectives: learning physical priors and injecting them through a richer motion representation. 
First, instead of relying on fully specified physics engines or purely semantic reasoning, we introduce a lightweight and learnable \textbf{Particle-based Graph Dynamics Simulator (PGDS)}, inspired by~\cite{sanchez2020learning}. 
By representing dynamic scenes as particle-based systems, PGDS learns physical evolution directly from data, providing generalizable and fine-grained priors for complex object interactions.
Second, we propose a \textbf{Visible–Invisible Motion Field (VIMF)} derived from a simulated 3D motion trajectory, which jointly models both observable motion in the reference view and physically emergent motion in future unseen regions. 
VIMF preserves richer 3D dynamic information and provides more complete guidance to the video generation model, enabling it to synthesize motions that are not only visually coherent but also physically plausible.
To support systematic evaluation, we further construct a benchmark for physics-aware video generation, covering diverse scenarios such as domino effects, free fall, collisions, and their compositional variants, with a total of 350 samples.  
Extensive experiments demonstrate that our method produces more accurate and stable motion trajectories, better preserves interaction consistency (e.g., contact and causality), and significantly reduces artifacts such as object interpenetration and unrealistic motion. 
Overall, our approach substantially improves physical plausibility compared to strong baselines.

Our contributions are summarized as follows:
\vspace{-2mm}
\begin{itemize}[leftmargin=1.0em,itemsep=-0.5pt]
    \item We introduce an efficient Particle-based Graph Dynamics Simulator to learn generalizable physical dynamics without manually specified physical parameters.
    \item We propose a Visible--Invisible Motion Field that jointly captures observable motion and physically predicted unseen motion, providing richer and more complete guidance for video generation.
    \item We build a benchmark for physics-aware video generation and demonstrate significant improvements in motion accuracy and physical plausibility over existing methods.
\end{itemize}

\section{Related Work}
\textbf{Physically Grounded Video Generation.}
Recent efforts toward physically grounded video generation can be broadly categorized into three directions. 
Early approaches~\citep{liu2024physgen,tan2024physmotion,chen2025physgen3d} rely on physics simulators to explicitly model object dynamics, including rigid body motion, deformation, and fluid behavior~\citep{tung2023physion++,Genesis}. 
While these methods provide structured physical signals, they are typically developed under controlled settings and depend on predefined scene representations. 
To improve scalability, subsequent works explore data-driven approximations of physical dynamics~\citep{yuan2025newtongen,ji2025physmaster,wang2025physctrl}. 
These methods relax the need for explicit simulation but often operate under constrained settings or specific scenario types, which limits their ability to generalize to complex multi-object interactions. 
More recently, a growing body of work~\citep{zhang2025think,bai2025moregen,arani2026physvid} leverages the physical commonsense knowledge embedded in large vision-language models (LVLMs)~\citep{wang2024qwen2,openai2024gpt4o} or pre-trained video encoders~\citep{zhang2025videorepa,ji2025physmaster} to guide video generation. 
Such approaches typically introduce intermediate reasoning processes, such as predicting key motion trajectories~\citep{gu2026place,yang2025vlipp,huang2025planning} or generating keyframes~\citep{huang2025vchain}, which are then used to guide video synthesis. 
In addition, some methods~\citep{cai2025phygdpo,zhang2025think} employ LVLMs as verifiers to assess physical consistency during or after generation (e.g., physical constraint checking or context verification). 
Despite these advances, existing approaches either rely on complex pipelines with limited scalability or provide coarse-grained guidance that lacks explicit modeling of underlying dynamics, making it challenging to enforce precise physical consistency.

\textbf{Controllable Video Generation with Motion Guidance.} 
Controlling motion in video generation has been extensively studied through various conditioning mechanisms. 
Some works~\citep{gillman2025force,narayanan2026phyco,gillman2026goal,geng2025motion} incorporate motion cues through additional control branches to provide flexible and intuitive control over motion patterns.
Another line of work integrates motion information directly into the generative process via attention-based mechanisms~\citep{gu2025diffusion,wang2023videocomposer,li2026spread}.  
In addition, some methods~\citep{arani2026physvid,chua2025wanmove} explore implicit control strategies where motion guidance is encoded in textual prompts or high-level instructions. 
Despite these advances, existing methods predominantly rely on 2D motion representations and view-dependent signals. 
As a result, they often fail to capture complete 3D dynamics, particularly for regions that are initially occluded or become visible over time, restricting their ability to model complex interactions and provide physically consistent guidance for video generation.

\textbf{Physical Dynamics Learning and Simulation.}
Modeling physical dynamics has been extensively studied in both classical simulation and learning-based frameworks. 
Traditional approaches rely on numerical solvers grounded in physical laws, such as finite element methods (FEM)~\citep{zienkiewicz1977finite}, position-based dynamics (PBD)~\citep{macklin2013position}, material point methods (MPM)~\citep{jiang2016material}, smoothed particle hydrodynamics (SPH)~\citep{peer2018implicit}, and mass-spring systems~\citep{liu2013fast}. 
These methods provide accurate and interpretable solutions but typically require detailed scene representations and carefully specified physical parameters, limiting their applicability in data-driven visual settings. 
To improve flexibility, neural approaches have been proposed to approximate physical processes, such as PINNs~\citep{raissi2019physics} and fluids~\citep{chu2022physics,wang2024physics}. 
Hybrid methods further integrate neural components into simulation pipelines to accelerate inference~\citep{feng2026elastogen}, though they often rely on structured representations or per-scene optimization. 
More recently, neural simulators based on graph neural networks have emerged as an effective paradigm for learning physical dynamics~\citep{sanchez2020learning} by modeling interactions between particles or objects.
Beyond these, recent works also explore diffusion models to model point-based physical dynamics~\citep{wang2025physctrl}, or neural Newtonian dynamics to predict low-dimensional latent states~\citep{yuan2025newtongen}.
While these approaches improve flexibility, their compact representations struggle to capture complex multi-object interactions; we therefore adopt a particle-based graph simulator for more expressive physical modeling.

\begin{figure}[!t]
    \centering
    \includegraphics[width=0.99\linewidth]{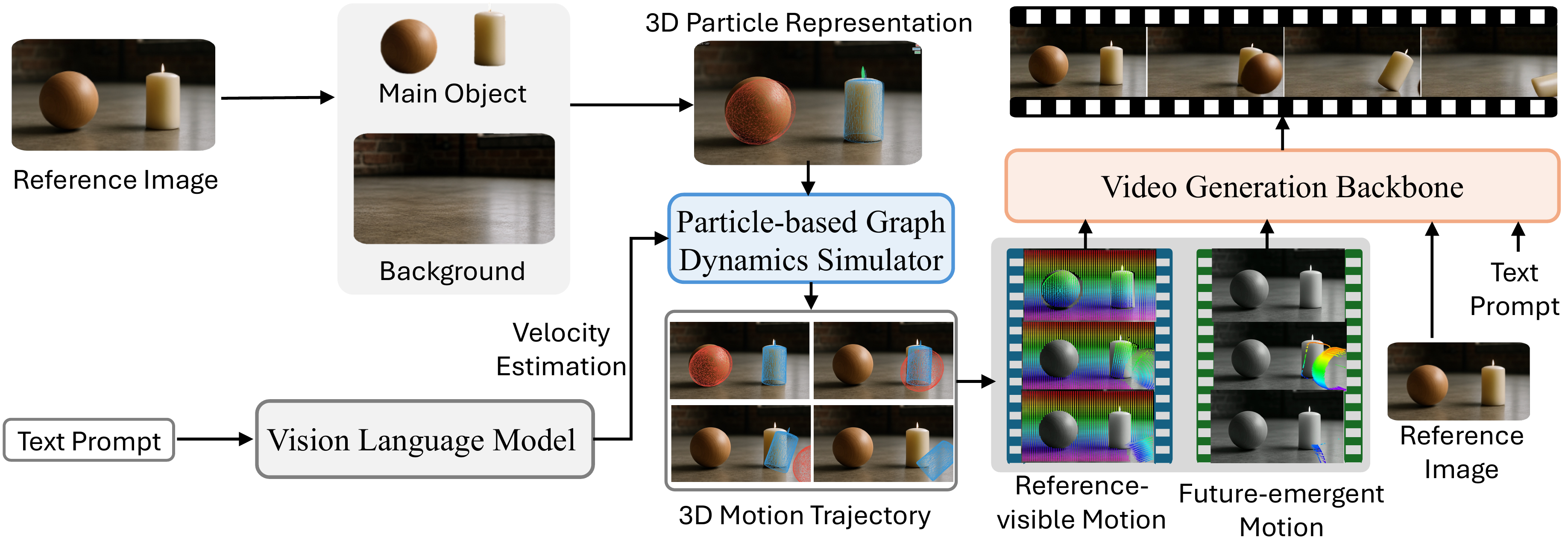}
    \vspace{-2mm}
    \caption{Overview of our proposed framework. Given an input image, we first lift it into a 3D scene particle representation. A particle-based graph dynamics simulator then predicts physically plausible 3D motion trajectories, which are rendered into reference-visible and future-emergent motions to guide physically consistent video generation.}
    \label{fig:method}
    \vspace{-3mm}
\end{figure}

\section{Method}

\vspace{-2mm}
\subsection{Overview}
\vspace{-2mm}

To generate physically consistent videos, we propose a unified framework that jointly addresses physical prior learning and seamless motion-guided video synthesis. 
As illustrated in Figure~\ref{fig:method}, given an input image $I$, we first lift it into a simulatable 3D scene representation. 
Then, we employ a particle-based graph dynamics simulator (PGDS) to model object interactions and predict physically plausible motion trajectories in 3D space (cf.~\S\ref{sec:dynamics}). 
The predicted 3D trajectories are then transformed into a unified motion representation that captures both observable motion and future motion in previously unseen regions in the given reference image. 
Finally, we inject this representation into a video generation model through a visible-invisible motion field (VIMF), enabling the synthesis of videos that are both visually coherent and physically plausible (cf.~\S\ref{sec:generation}).

\vspace{-2mm}
\subsection{Learning Physical Dynamics with Graph-Based Simulation}
\label{sec:dynamics}

\vspace{-1mm}
\paragraph{Input Image Lifting.}
Given an input image $I \in \mathbb{R}^{H\times W \times 3}$, we first lift it into a simulatable 3D scene representation. 
Specifically, we begin by estimating a pixel-wise depth map and converting it into a metric 3D point map $\mathcal{B} = \{\bm{p}_i\}_{i=1}^{N}$, where $p_i \in \mathbb{R}^{3}$ denotes the $i$-th 3D position. 
We then segment foreground objects using SAM3~\citep{carion2025sam3}, obtaining instance-level binary masks. 
For each segmented object, we perform single-image 3D reconstruction using SAM 3D~\citep{chen2025sam3d}, conditioned on the predicted point map. 
This step yields object-level 3D representations with metric poses, encoding their positions and scales relative to the camera. 
We then transform all objects into a unified scene coordinate system and estimate the ground plane via RANSAC-based fitting~\citep{miraldo2022unified,zhong2025plane}.
Finally, we aggregate all objects into a multi-object scene representation, which is then converted into a set of simulation particles, which serves as the input to our graph-based dynamics simulator.

\begin{figure}[!t]
    \centering
    \includegraphics[width=0.99\linewidth]{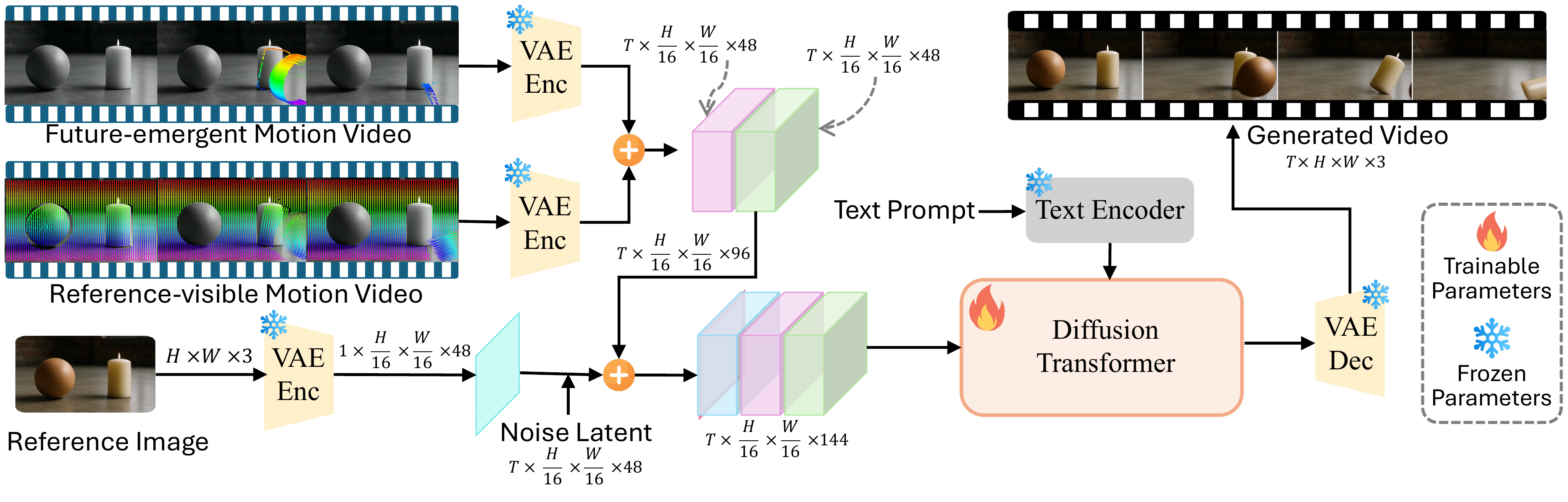}
    \vspace{-2mm}
    \caption{Architecture of the physics-guided video generation model. The visible and future visible motion fields are encoded by a frozen 3D VAE and concatenated with the noisy video latent along the channel dimension. A zero-initialized injection strategy preserves the pretrained model behavior while progressively incorporating motion guidance.}
    \label{fig:finetuned_model}
    \vspace{-2mm}
\end{figure}

\vspace{-1mm}
\paragraph{Particle-based Graph Dynamics Simulator.} 
Given simulation particles from the lifted 3D scene, we construct a graph $\mathcal{G}=(\mathcal{V},\mathcal{E})$, where nodes represent particles and edges connect neighbors within a radius $r$. 
Each particle is associated with a feature $\mathbf{z}_i = [\mathbf{c}_i;\mathbf{o}_i;\mathbf{s}_i]$, encoding material attributes, object-relative geometry, and dynamic state. 
Node and edge features are encoded as $\mathbf{h}_i^v = \phi^v(\mathbf{z}_i)$ and $\mathbf{h}_{ij}^e = \phi^e([\mathbf{h}_i^v;\mathbf{h}_j^v;\mathbf{r}_{ij};\|\mathbf{r}_{ij}\|])$, where $\mathbf{r}_{ij} = \mathbf{x}_i - \mathbf{x}_j$.
We perform $T$ rounds of message passing to model particle interactions:
\begin{equation}
\mathbf{m}_{ij}^{(t)} = \psi^e([\mathbf{h}_{ij}^e;\mathbf{e}_i^{(t)};\mathbf{e}_j^{(t)}]), \quad
\mathbf{e}_i^{(t+1)} = \mathbf{e}_i^{(t)} + \psi^v\!\left([\mathbf{h}_i^v; \sum_{j\in\mathcal{N}(i)} \mathbf{m}_{ij}^{(t)}]\right),
\end{equation}
where $\mathcal{N}(i)$ denotes the neighborhood of particle $i$, and $\mathbf{e}_i^{(0)}=\mathbf{0}$. 
Finally, particle velocities are predicted as $\hat{\mathbf{v}}_i = \phi^{\mathrm{dec}}(\mathbf{e}_i^{(T)})$, which are used to update positions and roll out trajectories.
This formulation enables PGDS to capture structured interactions and learn fine-grained physical dynamics in a particle-based representation.

\paragraph{Training Objective.}
We train the simulator by minimizing the mean squared error between predicted and ground-truth particle velocities. 
Given $\hat{\mathbf{v}}_i$ and $\mathbf{v}_i^*$, the loss is defined as 
\begin{equation}
\mathcal{L}_{\text{PGDS}} = \frac{1}{B} \sum_{g=1}^{B} \frac{1}{N_g} \sum_{i \in \mathcal{G}} \frac{1}{3}\left\lVert \hat{\mathbf{v}}_i - \mathbf{v}_i^* \right\rVert_2^2,
\end{equation}
where $N_g$ is the number of particles in graph $\mathcal{G}$ and $B$ is the batch size. 
This formulation ensures equal contribution from each scene regardless of its size.

\vspace{-1mm}
\paragraph{Simulation Capability and Generalization}
PGDS models local particle-level interactions rather than object-specific global geometry, which enables it to generalize across unseen object shapes and spatial configurations. 
Its material space is also extensible: new material types can be incorporated by adding corresponding simulated data and specifying them through the material attribute vector $\mathbf{c}_i$. 
Prior studies~\citep{tung2023physion++} further show that such a simulator can remain reliable even when the number of particles increases substantially at test time, suggesting strong scalability to more complex scenes. 
These properties make PGDS well-suited for open-ended video generation, where object geometry, particle density, and interaction patterns may differ from those observed during training.

\subsection{Physics-grounded Image-to-Video Generation}
\label{sec:generation}

\paragraph{Visible--Invisible Motion Field.}
Existing motion-guided video generation methods typically convert predicted motion into 2D optical flow or sparse point tracks. 
Although effective for conveying visible motion, such representations discard 3D spatial structure and only describe regions observable from the reference view. 
They therefore provide limited guidance for occluded surfaces or objects that become visible in later frames.
To address this limitation, we decompose physical motion into two complementary tracking signals. 
The first is a \emph{reference-visible motion tracking video} $\mathcal{M}_{\mathrm{vis}}$, which records the motion of points visible in the input frame. 
The second is a \emph{future-emergent motion tracking video} $\mathcal{M}_{\mathrm{fv}}$, which captures points that are initially occluded or outside the reference view but become visible in future frames. 
Together, these two signals form a visible--invisible motion representation that preserves both immediately observable motion and physically predicted future motion. 
More details are provided in \textcolor{black}{Appendix~\ref{app:vimf}}.
During training, we obtain these tracking signals from videos using a learning-based tracking pipeline. 
During inference, we derive them by projecting the simulated 3D trajectories from PGDS into the same tracking-video format.

\vspace{-1mm}
\paragraph{Unified Motion Encoding and Injection}
We build our physics-guided generator on top of Wan2.2-TI2V-5B~\citep{wan2025wan}. 
To inject our dual motion guidance, we encode $\mathcal{M}_{\mathrm{vis}}$ and $\mathcal{M}_{\mathrm{fv}}$ using the same frozen 3D VAE to yield the latent representations $\mathbf{z}_{\mathrm{vis}}$ and $\mathbf{z}_{\mathrm{fv}}$, and then concatenate them with the noisy video latent along the channel dimension:
\begin{equation}
\mathbf{z}_t =
\mathrm{Concat}\left[
\mathbf{z}_t,\,
\mathbf{z}_{\mathrm{vis}},\,
\mathbf{z}_{\mathrm{fv}}
\right].
\end{equation}
To preserve the pretrained model behavior, we adopt a zero-initialized injection strategy to allow the model to incorporate motion guidance progressively without disrupting the original generation capability. 
For appearance conditioning, we follow the standard TI2V scheme by replacing the first-frame latent with a clean reference, ensuring consistent visual grounding.

\vspace{-1mm}
\paragraph{Training Objective for Physics-Guided Generation.}
Following~\citep{lipman2022flow,esser2024scaling,wan2025wan}, we adopt the flow matching framework. 
Given a clean video latent $\mathbf{z}_0$ and Gaussian noise $\boldsymbol{\epsilon} \sim \mathcal{N}(\mathbf{0}, \mathbf{I})$, the noisy latent at timestep $t$ is:
\begin{equation}
    \mathbf{z}_t = (1 - \sigma_t)\,\mathbf{z}_0 + \sigma_t\,\boldsymbol{\epsilon},
    \label{eq:flow_matching}
\end{equation}
where $\sigma_t$ follows the noise schedule and $t$ is uniformly sampled. 
The model predicts the velocity field $\boldsymbol{\epsilon}-\mathbf{z}_0$ conditioned on text, the reference image, and the dual tracking latents:
\begin{equation}
\mathcal{L}_{\mathrm{gen}} =
\mathbb{E}_{\mathbf{z}_0,\boldsymbol{\epsilon},t}
\left[
w(\sigma_t)
\left\|
f_\theta(
\mathbf{z}^{\mathrm{in}}_t,
t,
\mathbf{c}_{\mathrm{text}}
)
-
(\boldsymbol{\epsilon}-\mathbf{z}_0)
\right\|_2^2
\right],
\end{equation}
where $w(\sigma_t)$ denotes the signal-to-noise weighting, $\mathbf{c}_{\mathrm{text}}$ is the text embedding, and $\mathbf{z}^{\mathrm{in}}_t$ contains the noisy video latent together with the visible and future-visible tracking latents. 
During training, the first-frame latent is kept clean and assigned zero timestep, indicating a noise-free visual reference.

\section{Settings}

\vspace{-1mm}
\paragraph{Dataset.}
For physical dynamics modeling, we train PGDS on the Physion++ dataset~\citep{tung2023physion++}, which contains approximately 16K samples generated by the ThreeDWorld simulator~\citep{gan2020threedworld} and covers eight types of physical interactions.
For motion-guided video generation, we combine simulated data with real-world videos from MiraData~\citep{ju2024miradata}. 
We further extract motion supervision using SpatialTracker2~\citep{xiao2025spatialtrackerv2}, thereby generating two trajectory sets: reference-visible and future-emergent trajectories. 
For evaluation, we focus on seven physics-aware interaction categories, including four primitive motion types (\textit{free fall}, \textit{domino}, \textit{collision}, and \textit{translation}) and three compositional scenarios (\textit{collision$+$domino}, \textit{free fall$+$domino}, and \textit{free fall$+$collision}). 
Each category contains 50 samples, each consisting of a reference image and a paired text prompt. 
In addition to single-object scenes, we explicitly include multi-object compositions (e.g., multiple objects or mixed categories), enabling evaluation under both simple and complex interaction settings.

\vspace{-1mm}
\paragraph{Implementation Details.}
PGDS operates on particle graphs with about 3000 nodes per scene, where edges connect particles within a radius $r{=}0.08$. 
Node and edge representations have latent dimensions of 200 and 300, respectively. 
We perform 10 rounds of message passing to model particle interactions. 
The model is trained for 1K epochs on four NVIDIA H200 GPUs with a global batch size of 32. 
We use the Adam optimizer with an initial learning rate of $10^{-4}$ and a {ReduceLROnPlateau} scheduler. 
To improve robustness, Gaussian noise with $\sigma = 3 \times 10^{-4}$ is injected into input velocities during training.
During the video generation training stage, we initialize the generation backbone from Wan2.2-TI2V-5B~\citep{wan2025wan}. 
To facilitate stable motion injection, we adopt an asymmetric learning rate scheme: the patch embedding layer is trained with a learning rate of $2 \times 10^{-4}$, while all other parameters use $2 \times 10^{-5}$. 
This design allows the newly introduced motion channels to adapt rapidly while preserving the pretrained generation capability. 
We optimize the model using AdamW~\citep{loshchilov2017decoupled} with $\beta_1=0.9$, $\beta_2=0.999$, weight decay $3 \times 10^{-2}$, and $\epsilon=10^{-10}$. 
Training uses a constant learning rate schedule with 100 warmup steps and gradient clipping at 0.5.

\vspace{-1mm}
\paragraph{Baselines and Evaluation metrics.}
We compare against two groups of methods. 
(1)~General video generation models (without trajectory injection): open-source models including Wan2.2-TI2V-5B~\citep{wan2025wan}, Wan2.2-I2V-A14B~\citep{wan2025wan}, HunyuanVideo-1.5 720p I2V~\citep{wu2025hunyuanvideo}, and CogVideoX1.5-5B-I2V~\citep{yang2024cogvideox}, as well as closed-source state-of-the-art models Wan2.7 and Veo3.1~\citep{google2026veo31}. 
(2)~Physics-guided generation models: PhysCtrl~\citep{wang2025physctrl}, NewtonGen~\citep{yuan2025newtongen}, and PhysT2V~\citep{xue2025phyt2v}. 
Following~\citep{bansal2024videophy}, we adopt two evaluation metrics. 
\textbf{(1) Semantic Adherence (SA):} measures the alignment between the generated video and the text prompt in content and motion (e.g., force direction and location); 
\textbf{(2) Physical Commonsense (PC):} evaluates whether the motion follows physically plausible dynamics consistent with the specified forces.

\begin{table}[t]\footnotesize
\vspace{-5px}
\caption{Quantitative results on seven physical video generation scenarios. Best results are \textbf{bold}. ``--'' indicates unsupported cases.}
\label{tab:results}
\centering
\resizebox{\textwidth}{!}{%
\begin{tabular}{l cc cc cc cc cc cc cc}
\toprule
\multirow{2}{*}{\textbf{Method}} & \multicolumn{2}{c}{\textbf{Domino}} & \multicolumn{2}{c}{\textbf{Collision}} & \multicolumn{2}{c}{\textbf{Free Fall}} & \multicolumn{2}{c}{\textbf{Coll.+Dom.}} & \multicolumn{2}{c}{\textbf{F.F.+Dom.}} & \multicolumn{2}{c}{\textbf{F.F.+Coll.}} & \multicolumn{2}{c}{\textbf{Translation}} \\
\cmidrule(lr){2-3}\cmidrule(lr){4-5}\cmidrule(lr){6-7}\cmidrule(lr){8-9}\cmidrule(lr){10-11}\cmidrule(lr){12-13}\cmidrule(lr){14-15}
 & SA & PC & SA & PC & SA & PC & SA & PC & SA & PC & SA & PC & SA & PC \\
\midrule
\rowcolor{groupgray}\multicolumn{15}{c}{\textit{\textbf{Open-source Video Generation Models}}} \\
Wan2.2-I2V-14B        & 2.483 & 3.017 & 2.750 & 3.083 & 2.317 & 3.033 & 2.217 & 3.050 & 2.533 & 2.950 & 2.517 & 3.000 & 3.000 & 4.000 \\
Wan2.2-TI2V-5B        & 2.650 & 3.033 & 2.917 & 3.067 & 2.183 & 3.083 & 2.200 & 3.033 & 2.467 & 3.033 & 2.183 & 3.083 & 3.000 & 4.100 \\
HunyuanVideo-1.5      & 2.550 & 3.017 & 2.867 & 3.067 & 2.317 & 3.017 & 2.233 & 3.000 & 2.617 & 2.950 & 2.467 & 3.000 & 3.050 & 3.933 \\
CogVideoX1.5-5B      & 2.450 & 3.200 & 2.833 & 3.017 & 2.333 & 3.000 & 2.300 & 2.983 & 2.650 & 3.000 & 2.333 & 3.000 & 3.050 & 3.933 \\
\midrule
\rowcolor{groupgray}\multicolumn{15}{c}{\textit{\textbf{Physics-aware Models}}} \\
PhysT2V               & 2.267 & 3.867 & 2.533 & 3.400 & 2.050 & 4.283 & 2.283 & 3.267 & 2.383 & 3.333 & 2.117 & 3.933 & 3.033 & 4.400 \\
NewtonGen             & -- & -- & -- & -- & -- & -- & -- & -- & -- & -- & -- & -- & 3.033 & 3.983 \\
PhysCtrl              & -- & -- & -- & -- & -- & -- & -- & -- & -- & -- & -- & -- & 2.867 & 4.567 \\
\midrule
\rowcolor{groupgray}\multicolumn{15}{c}{\textit{\textbf{Closed-source Video Generation Models}}} \\
Wan-2.7               & 2.200 & 2.800 & 2.800 & 3.000 & 2.400 & 3.000 & 2.200 & 3.000 & 2.400 & 2.800 & 2.200 & 3.000 & 3.000 & \textbf{4.600} \\
Veo 3.1               & 2.800 & 3.000 & 2.800 & 3.000 & 2.400 & 3.000 & 2.400 & 3.000 & 2.800 & 3.000 & 2.200 & 3.000 & 3.200 & 4.000 \\
\midrule
Ours                  & \textbf{3.976} & \textbf{4.013} & \textbf{4.160} & \textbf{4.394} & \textbf{4.347} & \textbf{4.328} & \textbf{3.854} & \textbf{4.016} & \textbf{3.985} & \textbf{4.017} & \textbf{4.327} & \textbf{4.589} & \textbf{4.013} & 4.578 \\
\bottomrule
\end{tabular}%
}
\vspace{-3px}
\end{table}

\begin{figure}[!t]
    \centering
    \includegraphics[width=0.99\linewidth]{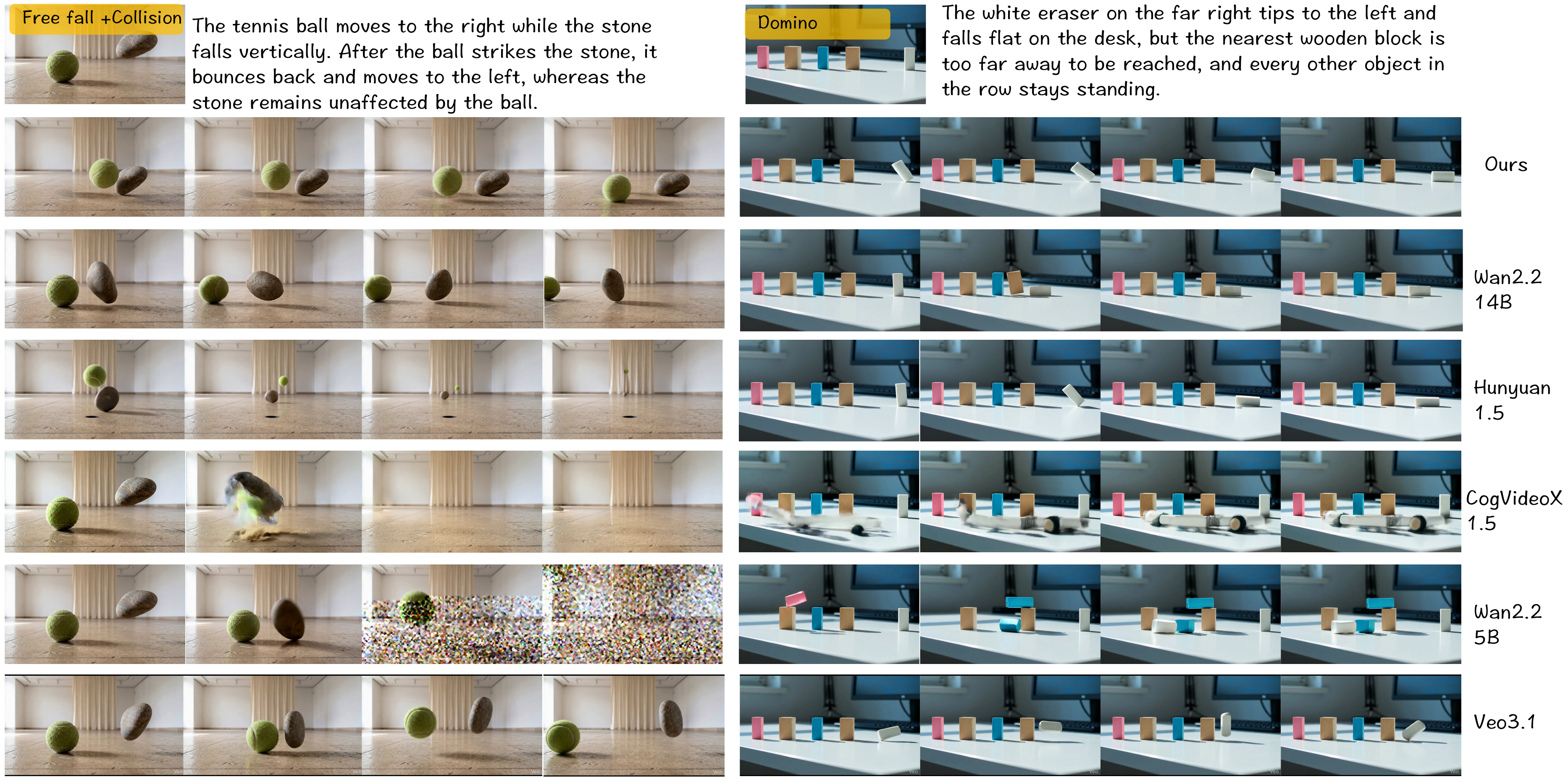}
    \caption{
    Qualitative comparison with existing methods.  Our method produces physically plausible dynamics with accurate object interactions (e.g., correct collision responses and domino chain propagation), while baseline models frequently exhibit artifacts such as object interpenetration, implausible motion trajectories, or failure to trigger expected physical events.
    }
    \label{fig:qualitative_comparison}
    \vspace{-4mm}
\end{figure}

\vspace{-2mm}
\section{Results and Discussion}
\vspace{-2mm}

\paragraph{Main Results on Physically Consistent Video Generation.}
We compare our method with open-source baselines, recent SOTA models, and prior physics-guided approaches on the proposed benchmarks. 
As shown in Table~\ref{tab:results}, our model achieves the best overall performance across most scenarios. 
Compared to its backbone Wan2.2-TI2V-5B, our model consistently improves both SA and PC scores, demonstrating the effectiveness of integrating learned physical dynamics and structured motion guidance.
We observe that most models perform well on the \textit{Translation} benchmark, which involves single-object motion without interactions. 
However, performance drops significantly as the task complexity increases, especially in compositional scenarios (e.g., \textit{collision+domino}, \textit{free fall+collision}). 
In contrast, our method maintains stable performance, indicating stronger capability in modeling multi-object interactions and causal dynamics.
All methods achieve relatively lower scores on domino-related tasks that involve dense object layouts and frequent occlusions, as illustrated in Figure~\ref{fig:qualitative_comparison}.
They often fail to propagate contact dynamics correctly (e.g., incomplete domino chains) or produce unstable and inconsistent object motions. 
In contrast, our method generates coherent interaction patterns and accurately models contact propagation, demonstrating stronger capability in capturing structured physical dynamics under complex multi-object settings.
Additional visualizations can be found in \textcolor{black}{Appendix~\ref{app:more_vis}}.

\begin{table}[!t]
\begin{minipage}[t]{0.35\linewidth}
\vspace{2pt}
\captionsetup{justification=raggedright,singlelinecheck=false,hypcap=false}
\captionof{table}{Results (best in \textbf{bold}) on the compatible \mbox{VideoPhy2} subset.}
\label{tab:videophy2}
\centering
\fontsize{8}{9}\selectfont
\setlength{\tabcolsep}{5pt}
\renewcommand{\arraystretch}{1.20}
\begin{tabular*}{\linewidth}{@{}l@{\extracolsep{\fill}}cc@{}}
\toprule
\textbf{Model} & \textbf{SA} $\uparrow$ & \textbf{PC} $\uparrow$ \\
\midrule
CogVideoX-5B        & 3.21 & 3.65 \\
Wan2.2-5B          & 3.14 & 3.46 \\
Ours (visible-only) & 3.47 & 3.78 \\
\midrule
Ours               & \textbf{4.01} & \textbf{4.13} \\
\bottomrule
\end{tabular*}
\end{minipage}%
\hspace{0.03\linewidth}%
\begin{minipage}[t]{0.62\linewidth}
\vspace{2pt}
\captionsetup{justification=RaggedRight,singlelinecheck=false,hypcap=false}
\captionof{table}{PGDS rollouts on 7 Physion++ scenarios. Position MSE: mean squared particle position error; CoM: center-of-mass error (\% object size); rotation: angular deviation.}
\label{tab:pgds_rollout}
\centering
\fontsize{8}{9}\selectfont
\setlength{\tabcolsep}{4pt}
\renewcommand{\arraystretch}{1.10}
\begin{tabular*}{\linewidth}{@{\extracolsep{\fill}}crrr@{}}
\toprule
\makecell{\textbf{Horizon}\\$k$} & \makecell{\textbf{Position}\\\textbf{MSE (cm$^2$)} $\downarrow$} & \makecell{\textbf{CoM error}\\\textbf{(\% size)} $\downarrow$} & \makecell{\textbf{Rotation}\\\textbf{error ($^{\circ}$)} $\downarrow$} \\
\midrule
30 & 0.7 & 0.5 & 1.3 \\
60 & 1.7 & 1.2 & 3.8 \\
90 & 3.0 & 2.6 & 7.5 \\
\bottomrule
\end{tabular*}
\par
\end{minipage}
\par
\vspace{\floatsep}
\centering
\caption{PGDS motion prediction on seven Physion++ test scenarios.
Vel. RMSE is the root mean squared particle velocity error over complete rollouts. Acc-V reports the percentage of particles meeting strict ($0.05$~m/s or $5\%$) or relaxed ($0.10$~m/s or $10\%$) velocity-error thresholds. Kinetic-energy bias is $(E_{\mathrm{pred}}-E_{\mathrm{GT}})/E_{\mathrm{GT}}\times100\%$.}
\label{tab:pgds_accuracy}
\label{tab:pgds_velocity}
\fontsize{8}{9}\selectfont
\setlength{\tabcolsep}{3pt}
\begin{tabularx}{\linewidth}{@{}l*{4}{>{\centering\arraybackslash}X}@{}}
\toprule
\textbf{Scenario} & \makecell{\textbf{Vel. RMSE}\\\textbf{(m/s)} $\downarrow$} & \makecell{\textbf{Acc-V}\\\textbf{Strict (\%)} $\uparrow$} & \makecell{\textbf{Acc-V}\\\textbf{Relaxed (\%)} $\uparrow$} & \makecell{\textbf{Energy}\\\textbf{Bias (\%)}} \\
\midrule
Dominoes & 0.0077 & 99.20 & 99.65 & \textminus{}4.7 \\
Collide  & 0.0359 & 95.04 & 98.79 & \textminus{}5.5 \\
Support  & 0.0209 & 96.16 & 98.36 & \textminus{}6.5 \\
Link     & 0.0284 & 94.51 & 98.14 & +2.6 \\
Contain  & 0.0171 & 96.59 & 98.94 & \textminus{}3.8 \\
Roll     & 0.0384 & 95.48 & 98.27 & \textminus{}5.5 \\
Drop     & 0.0264 & 94.48 & 96.47 & \textminus{}1.2 \\
\midrule
\textbf{Average} & \textbf{0.0250} & \textbf{95.92} & \textbf{98.37} & \textbf{\textminus{}3.5} \\
\bottomrule
\end{tabularx}
\vspace{-2mm}
\end{table}

We further evaluate PhysT2V, NewtonGen, and PhysCtrl. 
NewtonGen and PhysCtrl are limited to simple motion settings and are only applicable to the Translation benchmark. 
PhysT2V achieves competitive PC scores, indicating its strength in improving physical plausibility via LLM-based refinement. 
However, its SA scores are generally lower, suggesting a trade-off between physical consistency and semantic fidelity.  
In contrast, our trajectory-based framework provides more stable and accurate guidance.

For external evaluation, we use \mbox{VideoPhy2}~\citep{bansal2025videophy} samples compatible with our setting.
\mbox{Table~\ref{tab:videophy2}} reports SA/PC scores of $4.01$/$4.13$ for our full model, the highest among all compared models.
The full model also improves both scores over the visible-only variant, supporting the effectiveness of future-emergent guidance.
Evaluation details and prompts are provided in Appendix~\ref{app:evaluation_protocol}.

\vspace{-2mm}
\paragraph{Prediction Accuracy of PGDS.}
As shown in Table~\ref{tab:pgds_accuracy}, PGDS achieves high motion prediction accuracy across all seven physical scenarios in Physion++.
Velocity errors remain low across scenarios, with an average strict Acc-V of $95.92\%$, indicating that the model accurately captures motion changes under diverse physical interactions.
The small discrepancy between predicted and ground-truth kinetic energy further indicates consistency with the ground-truth dynamics.

Although position, center-of-mass, and rotation errors accumulate with the prediction horizon (Table~\ref{tab:pgds_rollout}), they remain small after 90 steps, indicating accurate translation and rotation predictions over multi-step rollouts.
These results show that PGDS produces accurate 3D motion trajectories across diverse physical scenarios, providing reliable physical guidance for downstream video generation.
We further evaluate downstream video generation under incomplete motion guidance and upstream perception errors, with detailed robustness experiments provided in Appendix~\ref{app:perception_robustness}.

\vspace{-2mm}
\paragraph{Physical Dynamics Simulation Efficacy.}
We compare the efficiency of different physical prior sources in Table~\ref{tab:efficiency}. 
Simulator-based methods (e.g., PhysGen, PhysMotion) require extensive manual design (5–10 parameters) and precise 3D meshes, resulting in high setup cost and runtime ($\sim$10 minutes), while showing limited generalization. 
Data-driven approaches reduce manual effort ($\sim$2–5 parameters) and runtime ($\sim$6–8 minutes), but their generalization remains limited or moderate. 
LVLM-based methods eliminate manual configuration and achieve the fastest runtime ($\sim$2 minutes), yet lack fine-grained physical modeling. 
In contrast, PGDS requires no manual parameter tuning, maintains a competitive runtime ($\sim$5 minutes), achieves strong generalization to unseen shapes, and provides a better balance between efficiency and physical modeling capability.
We also report the end-to-end runtime and peak GPU memory usage of our full model, including stage-wise measurements and conditioning overhead, in Appendix~\ref{app:runtime_memory}.

\begin{table}[t]
\centering
\caption{
Configuration efficiency of trajectory generation methods. We compare setup cost and generalization with time and parameter estimates from public implementations.
}
\label{tab:efficiency}
\fontsize{8}{8}\selectfont
\setlength{\tabcolsep}{3.0mm}
\begin{tabular}{lccccc}
\toprule
\textbf{Method} & \textbf{Type} & \textbf{Manual Phys.} & \textbf{Precise 3D} & \textbf{Total Runtime} & \textbf{Unseen} \\
 & & \textbf{Params} $\downarrow$ & \textbf{Mesh} & \textbf{(incl.\ manual tuning)} $\downarrow$ & \textbf{Shapes} \\
\midrule
PhysGen~\cite{liu2024physgen}        & Simulator    & $\sim$6--10 & Required     & $\sim$10 min & Limited  \\
PhysMotion~\cite{tan2024physmotion}  & Simulator    & $\sim$6--8  & Required     & $\sim$10 min & Limited  \\
PhysGen3D~\cite{chen2025physgen3d}   & Simulator    & $\sim$5--7  & Required     & $\sim$10 min & Limited  \\
\midrule
PhysCtrl~\cite{wang2025physctrl}     & Data-driven  & $\sim$3--5  & Partial      & $\sim$8 min  & Moderate \\
NewtonGen~\cite{yuan2025newtongen}   & Data-driven  & $\sim$2--3  & Not required & $\sim$6 min  & Limited  \\
\midrule
PhysT2V~\cite{xue2025phyt2v}         & LVLM-based   & --          & Not required & $\sim$2 min  & Moderate \\
\midrule
\textbf{PGDS (Ours)}                 & Learned Sim. & \textbf{0}  & Required & \textbf{$\sim$5 min} & \textbf{Strong} \\
\bottomrule
\end{tabular}%
\vspace{-2mm}
\end{table}

\begin{figure}
    \centering
    \includegraphics[width=0.99\linewidth]{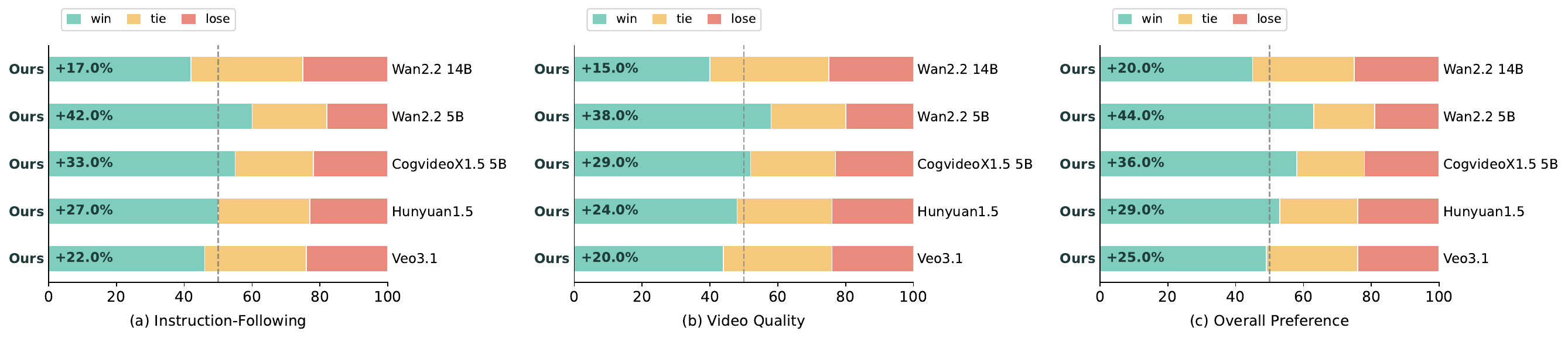}
    \caption{Pairwise human evaluation comparing our method against baseline models. For each comparison, human annotators judge which method produces better results across three criteria: (a) instruction-following, measuring how faithfully the generated video adheres to the given text prompt; (b) video quality, assessing visual fidelity, temporal coherence, and artifact severity; and (c) overall preference. Results are reported as win/tie/lose rates (\%).}
    \label{fig:human_eval}
    \vspace{-2mm}
\end{figure}

\paragraph{Human Evaluation}
We conduct pairwise human evaluation, with results shown in Figure~\ref{fig:human_eval}. 
Our method consistently outperforms all baselines across all three criteria.
In instruction following, our model achieves clear gains (e.g., +42.0\% over Wan2.2-5B), indicating better alignment with the input prompt. 
For video quality, we observe similar improvements (e.g., +38.0\%), suggesting stronger visual fidelity and temporal consistency. 
These advantages lead to higher overall preference (up to +44.0\%), demonstrating that our method produces more desirable results.
Overall, the results confirm that our approach improves both physical plausibility and perceptual quality.

\begin{table}[!t]
\centering
\vspace{-5px}
\caption{Comparison of physical prior injection strategies and the visible-only ablation across four scenarios. Our visible-only and full variants use the same diffusion backbone; the full variant uses both reference-visible and future-emergent tracking videos. Best results are \textbf{bold}.}
\label{tab:ablation_injection}
\fontsize{8}{8}\selectfont
\setlength{\tabcolsep}{3.4mm}
\begin{tabular}{l cc cc cc cc}
\toprule
\multirow{2}{*}{\textbf{Method}} & \multicolumn{2}{c}{\textbf{Domino}} & \multicolumn{2}{c}{\textbf{Collision}} & \multicolumn{2}{c}{\textbf{Free Fall}} & \multicolumn{2}{c}{\textbf{Translation}} \\
\cmidrule(lr){2-3}\cmidrule(lr){4-5}\cmidrule(lr){6-7}\cmidrule(lr){8-9}
 & SA & PC & SA & PC & SA & PC & SA & PC \\
\midrule
Go with Flow~\citep{burgert2025go}      & 2.583 & 3.389 & 3.275 & 3.691 & 3.670 & 3.928 & 3.568 & 4.419 \\
Wan-Move~\citep{chua2025wanmove}              & 2.672 & 3.351 & 3.594 & 3.771 & 3.892 & 4.017 & \textbf{4.167} & \textbf{4.602} \\
Diffusion as Shader~\citep{gu2025diffusion}   & 2.640 & 3.295 & 3.697 & 3.829 & 3.764 & 3.923 & 3.821 & 4.493 \\
\midrule
Ours (visible-only)   & 2.693 & 3.348 & 3.693 & 3.717 & 3.972 & 4.004 & 3.837 & 4.521 \\
Ours (full)           & \textbf{3.976} & \textbf{4.013} & \textbf{4.160} & \textbf{4.394} & \textbf{4.347} & \textbf{4.328} & 4.013 & 4.578 \\
\bottomrule
\end{tabular}
\vspace{-5px}
\end{table}

\begin{figure}[!t]
    \centering
    \includegraphics[width=0.89\linewidth]{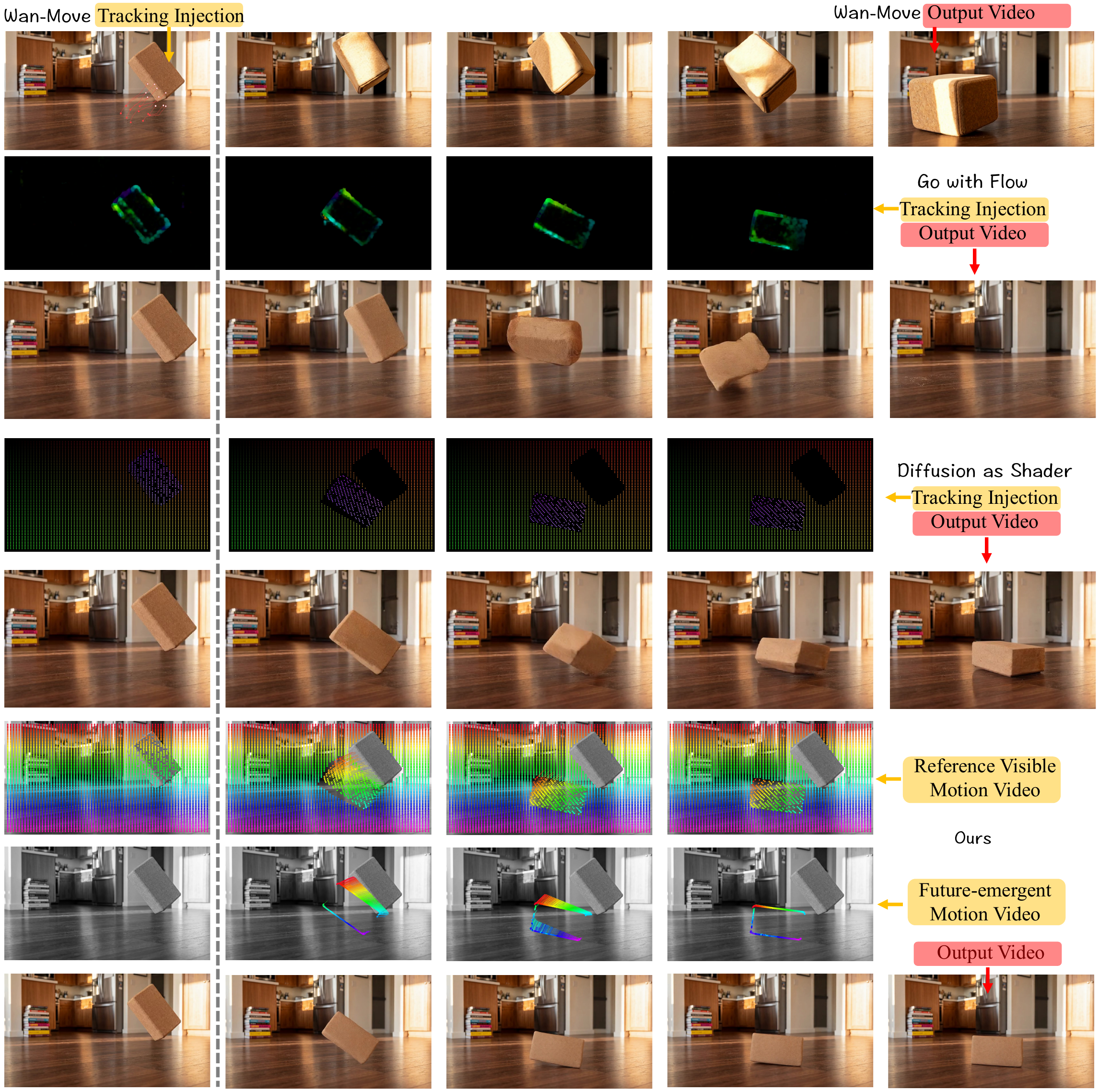}
    \vspace{-1mm}
    \caption{Comparison with different physical prior injection methods.}
    \label{fig:motion-comparison}
    \vspace{-4mm}
\end{figure}

\vspace{-2mm}
\paragraph{Comparison of Physical Prior Injections.}
We compare physical prior injection strategies by converting the same PGDS-predicted trajectories into different motion representations and feeding them to existing models.
Specifically, we provide optical flow to Go with Flow, 10 sparse tracking points to Wan-Move, and visible tracking videos to Diffusion as Shader, while retaining the original pretrained weights of each method.
As shown in Table~\ref{tab:ablation_injection}, our full model achieves the highest SA and PC scores on Domino, Collision, and Free Fall.
The optical flow used by Go with Flow struggles to fully represent object motion under occlusions and complex interactions, leading to motion errors and generation artifacts during interactions, as illustrated in Figure~\ref{fig:motion-comparison}.
Diffusion as Shader relies solely on visible trajectories and may lack motion information for newly revealed regions when objects rotate or become occluded, limiting the effectiveness of its guidance.
Wan-Move, which uses the Wan2.2-I2V-A14B backbone, performs well on Translation but provides less accurate motion control than our full model in complex interaction scenarios.

To further isolate the contribution of future visible motion, we conduct an ablation using the same diffusion backbone, comparing visible-only tracking videos with the joint use of reference-visible and future visible tracking videos.
The visible-only variant yields lower SA and PC scores than the full model in all four scenarios, with particularly pronounced gains from dual-stream guidance on Domino and Collision.
These results indicate that motion visible from the reference view alone is insufficient to fully describe occlusions and object interactions, whereas future visible trajectories supply motion information for regions that become visible later.
Combining the two types of trajectories provides more complete physical guidance for video generation, improving both semantic adherence and physical plausibility.

\vspace{-2mm}
\section{Conclusion}
\vspace{-2mm}
In this work, we propose \textbf{Seeing the Unseen}, a unified framework that combines a particle-based graph dynamics simulator with visibility-aware trajectory guidance to enhance physically grounded video generation. 
By explicitly modeling both observable motion and future-emergent dynamics, our method provides more complete and structured physical priors for video synthesis.
Extensive experiments demonstrate that our approach consistently outperforms existing methods across diverse challenging physical interaction scenarios, particularly in complex multi-object settings. 
Both quantitative results and human evaluations validate that our method improves physical plausibility, interaction consistency, and perceptual quality.
Overall, our results highlight the importance of integrating structured physical modeling and visibility-aware motion representations. 
We hope this work provides a step toward bridging video generation and physical reasoning and inspires future research on physics-grounded generative models.

\bibliographystyle{plainnat}
\bibliography{reference}

\appendix

\section{Ethic Statement}
This work focuses on improving the physical consistency of video generation models. 
We do not introduce new datasets involving human subjects or sensitive personal information, and all training data are derived from publicly available or synthetic sources.
Potential risks may arise from the misuse of video generation technology, such as generating misleading or fabricated content. 
While our method improves physical plausibility, it does not aim to create deceptive content, and we encourage responsible use aligned with existing guidelines for generative models.
Overall, we believe our work contributes positively to advancing reliable and physically grounded video generation, which can benefit applications such as simulation, education, and scientific visualization.

\section{LLM Usage}

Large language models (LLMs) were used solely for language polishing and improving the clarity of writing. 
All technical content, including model design, experiments, and analysis, was developed by the authors. 
The use of LLMs does not affect the originality or validity of the research contributions.

\section{Limitations}
Despite its effectiveness, our method is currently evaluated on a limited set of physical scenarios, primarily focusing on structured interactions such as collisions, domino effects, and free-fall motion. 
While the proposed framework is general and can be extended to more complex physical processes (e.g., fluid dynamics or highly deformable objects), such extensions require additional data and simulation design.
Moreover, our approach relies on an intermediate 3D scene lifting stage, which may introduce errors in geometry estimation that can propagate to downstream dynamics simulation and video generation.
We leave the exploration of broader physical domains and more robust 3D perception as future work.

\section{Detailed Method}

\subsection{Particle-based Graph Dynamics Simulator}
\begin{figure}[!t]
\centering
\includegraphics[width=0.99\linewidth]{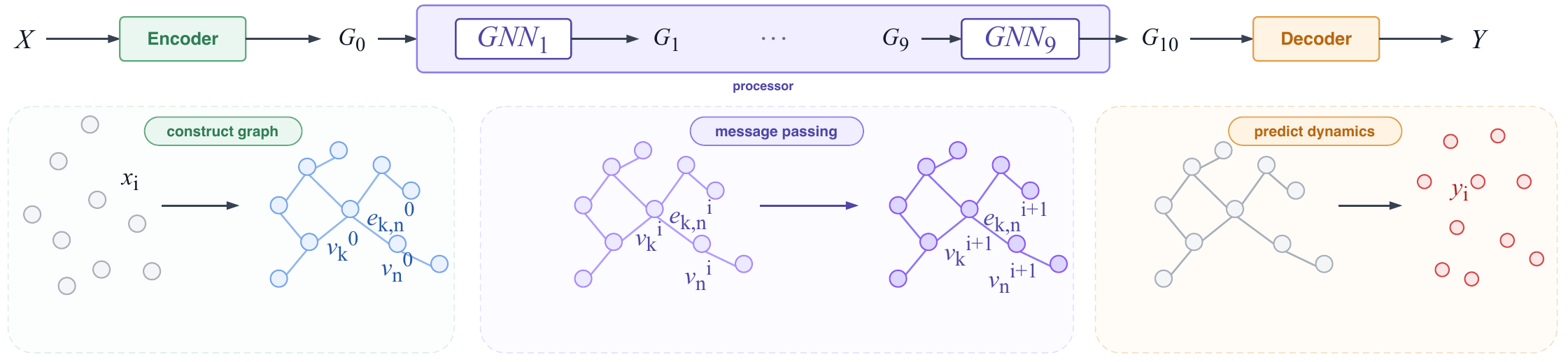}
\vspace{-2mm}
\caption{Architecture of the Particle-based Graph Dynamics Simulator (PGDS).
The simulator follows an Encode--Process--Decode pipeline.
Given particle states, a graph is first constructed based on spatial proximity.
Node and edge features are encoded to capture material properties and geometric relations.
A message-passing processor then iteratively propagates interaction effects across the graph.
Finally, a decoder predicts particle velocities, which are used to update positions and roll out future dynamics.
}
    \label{fig:app_gns}
\end{figure}

The Particle-based Graph Dynamics Simulator (PGDS) predicts particle velocities through iterative message passing over a graph constructed from the simulated scene. 
As illustrated in Figure~\ref{fig:app_gns}, the model follows an Encode--Process--Decode architecture, consisting of a particle encoder, a relation encoder, a message-passing processor, and a particle predictor.

\textbf{Encoding.}
Each particle $i$ is associated with an input feature vector
$\mathbf{z}_i = [\mathbf{c}_i; \mathbf{o}_i; \mathbf{s}_i]$,
where $\mathbf{c}_i \in \mathbb{R}^C$ is a one-hot material attribute, $\mathbf{o}_i \in \mathbb{R}^6$ encodes the object-relative offset, and $\mathbf{s}_i \in \mathbb{R}^6$ denotes the (masked) velocity state. 
The particle encoder $\phi^v$ maps this feature to a latent representation:
\begin{equation}
\mathbf{h}_i^v = \phi^v(\mathbf{z}_i).
\end{equation}

The offset $\mathbf{o}_i = \mathbf{x}_i - \bar{\mathbf{x}}_{g(i)}$ is defined relative to the centroid of object $g(i)$.
For rigid bodies, this encodes structural constraints by preserving relative positions; for fluid particles, $\mathbf{o}_i$ is set to zero to allow free motion.

For each edge $e_{ij}$ connecting neighboring particles, we compute a relation embedding based on relative geometry:
\begin{equation}
\mathbf{h}_{ij}^e =
\phi^e\left([\mathbf{h}_i^v; \mathbf{h}_j^v; \mathbf{r}_{ij}; \|\mathbf{r}_{ij}\|]\right),
\end{equation}
where $\mathbf{r}_{ij} = \mathbf{x}_i - \mathbf{x}_j$ denotes the relative displacement.

\textbf{Processing.}
The processor performs $T$ rounds of message passing (with $T=10$ in our implementation) to propagate interaction effects across particles. 
Let $\mathbf{e}_i^{(t)}$ denote the particle effect at iteration $t$, initialized as $\mathbf{0}$. 
At each iteration, edge messages and particle updates are computed as:
\begin{align}
\mathbf{m}_{ij}^{(t)} &= 
\psi^e\left([\mathbf{h}_{ij}^e; \mathbf{e}_i^{(t)}; \mathbf{e}_j^{(t)}]\right), \\
\mathbf{e}_i^{(t+1)} &=
\mathbf{e}_i^{(t)} +
\psi^v\left([\mathbf{h}_i^v; \sum_{j \in \mathcal{N}(i)} \mathbf{m}_{ij}^{(t)}]\right),
\end{align}
where $\mathcal{N}(i)$ denotes the neighbors of particle $i$, and the residual connection stabilizes iterative updates.

\textbf{Decoding.}
After $T$ iterations, the final particle effect $\mathbf{e}_i^{(T)}$ is decoded into a velocity prediction:
\begin{equation}
\hat{\mathbf{v}}_i = \phi^{\mathrm{dec}}(\mathbf{e}_i^{(T)}),
\end{equation}
where $\phi^{\mathrm{dec}}$ is a multi-layer perceptron. 
The predicted velocities are then used to update particle positions and roll out future trajectories.

\subsection{Unified Motion Encoding and Injection}
\label{sec:app_motion_injection}

\begin{figure}[!t]
    \centering
    \includegraphics[width=0.99\linewidth]{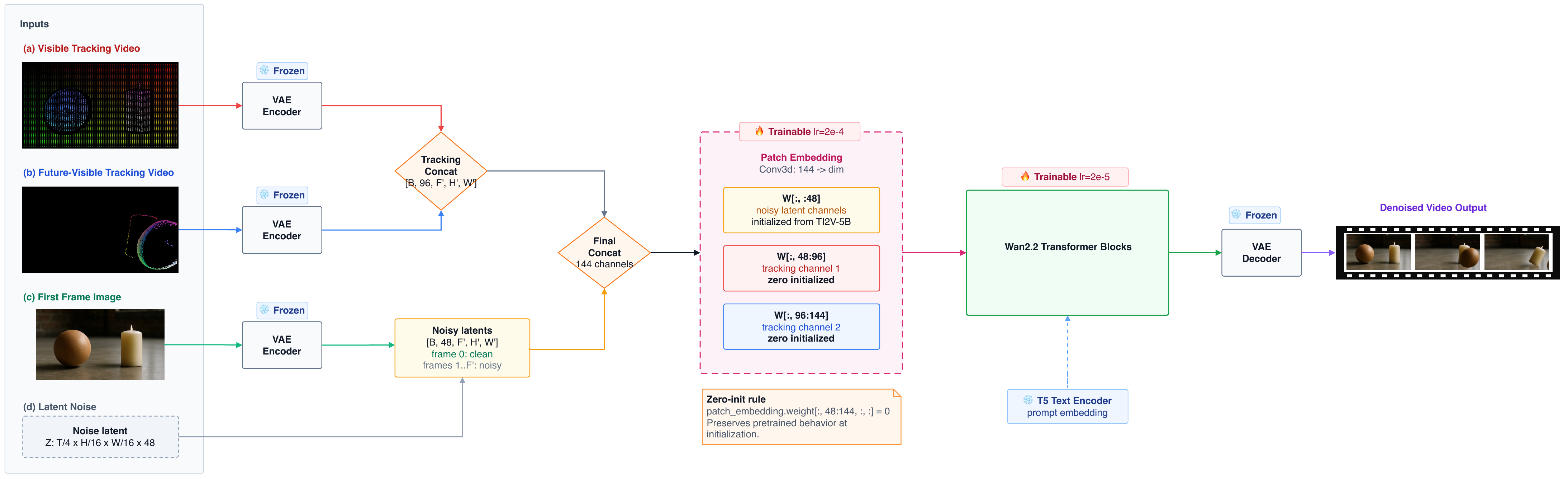}
    \caption{\textbf{Unified motion encoding and injection pipeline.}
The visible tracking video $\mathcal{V}_{\mathrm{vis}}$ and future-visible tracking video $\mathcal{V}_{\mathrm{fv}}$ are encoded using a frozen 3D VAE to obtain two latent sequences. 
These are concatenated with the noisy video latent to form a multi-channel input to the diffusion transformer. 
To preserve pretrained behavior, the patch embedding weights corresponding to tracking channels are zero-initialized, while the original weights are copied from the pretrained model. 
The first frame is injected via noise replacement to provide appearance conditioning.
}
    \label{fig:app_motion_injection}
\end{figure}

We provide a detailed implementation of motion injection in Figure~\ref{fig:app_motion_injection}. 
Our model is built upon Wan2.2-TI2V-5B, a text-and-image-to-video diffusion transformer with a 3D VAE encoder and a DiT backbone.

\textbf{Reference frame conditioning.}
We follow the standard TI2V noise-replacement scheme to incorporate appearance information. 
Specifically, the first frame is encoded into a clean latent $\mathbf{z}_0^{\mathrm{ref}}$ and replaces the noisy latent at the initial temporal position:
\begin{equation}
\mathbf{z}_t^{(0)} \leftarrow \mathbf{z}_0^{\mathrm{ref}}.
\end{equation}
This mechanism injects appearance conditioning without introducing additional input channels and keeps motion guidance disentangled from visual content.

\textbf{Latent construction.}
In the original model, the denoising network operates on a 48-channel latent produced by a 3D causal VAE with $16\times$ spatial and $4\times$ temporal compression. 
To incorporate motion guidance, we encode both the reference-visible tracking video $\mathcal{V}_{\mathrm{vis}}$ and the future-emergent tracking video $\mathcal{V}_{\mathrm{fv}}$ using the same frozen VAE, obtaining two additional 48-channel latent sequences:
\begin{equation}
\mathbf{z}_t = \mathrm{Concat}\left[
\mathbf{z}_t,\,
\mathbf{z}_{\mathrm{vis}},\,
\mathbf{z}_{\mathrm{fv}}
\right],
\end{equation}
where $\mathbf{z}_t$ denotes the noisy video latent. 
This results in a 144-channel tensor that is tokenized by the original 3D patch embedding layer.

\textbf{Zero-initialized motion injection.}
Expanding the input channels of the patch embedding introduces a distribution mismatch with the pretrained backbone. 
To mitigate this, we adopt a zero-initialization strategy: weights corresponding to the original 48 channels are copied from the pretrained checkpoint, while weights for the two tracking channels are initialized to zero. 
Formally, let $\mathbf{W} = [\mathbf{W}_{\text{orig}}, \mathbf{W}_{\text{track}}]$ denote the convolutional kernel, where
\begin{equation}
   \mathbf{W}_{\text{track}} = \mathbf{0}. 
\end{equation}
This ensures that the injected motion signals do not affect the network at initialization, allowing the model to gradually learn to utilize motion guidance during fine-tuning.

\section{Extended Settings}
\subsection{Automatic Evaluation Protocol}
\label{app:evaluation_protocol}
We use the open-source VideoPhy2-AutoEval-7B model~\citep{bansal2025videophy} to evaluate Semantic Adherence (SA) and Physical Commonsense (PC).
Both metrics are scored from 1 to 5, with higher scores indicating better performance.
We use the official VideoPhy2 evaluation prompts reproduced below, where \texttt{[V]} denotes the video being evaluated and \texttt{[T]} denotes its corresponding text description.

\paragraph{Semantic Adherence (SA).}
\begin{quote}
\small
[V] Does this video match the description: `[T]'? Please rate the video on a scale from 1 to 5, where 5 indicates a perfect match and 1 indicates no relevance.
\end{quote}

\paragraph{Physical Commonsense (PC).}
\begin{quote}
\small
[V] Does this video adhere to physical laws? Rate the video on a scale from 1 to 5, where 5 means full compliance and 1 means significant violations.
\end{quote}

\subsection{End-to-End Runtime and Memory}
\label{app:runtime_memory}
All runtime and GPU memory measurements in this subsection are performed on a single NVIDIA H200 GPU.
Table~\ref{tab:end_to_end_runtime} reports the stage-wise runtime and peak GPU memory usage for scenes with 1, 3, and 5 foreground objects.
The full pipeline consists of 3D lifting (S1), mesh-to-particle conversion and PGDS simulation (S2), dual-stream tracking rendering (S3), and video diffusion generation (S4).
The end-to-end runtimes are $181.9$, $189.0$, and $197.6$ seconds, respectively, with an overall peak GPU memory usage of $40.8$~GiB.
The stages preceding video diffusion (S1--S3) require approximately $30$--$50$ additional seconds.

\begin{table}[htbp]
\centering
\caption{End-to-end runtime and peak GPU memory usage on a single NVIDIA H200 GPU. The total covers stages S1--S4.}
\label{tab:end_to_end_runtime}
\setlength{\tabcolsep}{3pt}
\begin{tabularx}{\linewidth}{@{}l>{\raggedright\arraybackslash}Xrrrl@{}}
\toprule
\textbf{Stage} & \textbf{Component} & \textbf{1 Object} & \textbf{3 Objects} & \textbf{5 Objects} & \textbf{Peak VRAM} \\
\midrule
S1 & 3D lifting (segmentation + monocular depth + single-image 3D reconstruction) & 10.1~s & 15.8~s & 24.7~s & 28.7--40.8~GiB \\
S2 & Mesh-to-particle conversion & 4.0~s & 4.9~s & 5.1~s & $<1.2$~GiB \\
S2 & PGDS simulation & 1.3~s & 1.3~s & 1.3~s & 526~MiB \\
S3 & VIMF dual-stream tracking rendering & 15.9~s & 16.0~s & 15.9~s & $<1.5$~GiB \\
S4 & Video diffusion generation (Ours) & 150.6~s & 151.0~s & 150.6~s & 39.67~GiB \\
\midrule
\multicolumn{2}{@{}l}{\textbf{End-to-end total}} & \textbf{181.9~s} & \textbf{189.0~s} & \textbf{197.6~s} & \textbf{40.8~GiB} \\
\bottomrule
\end{tabularx}
\end{table}

To isolate the additional cost of dual-stream conditioning, we compare the video-generation stage with the base TI2V-5B backbone in Table~\ref{tab:conditioning_overhead}.
The input latent channels increase from 48 to 144, while the runtime of 35 denoising steps increases by only $0.20$ seconds and peak GPU memory increases by $84$~MiB.
Encoding the two tracking videos requires $5.95$ seconds.
The total video-generation time increases from $144.20$ to $150.28$ seconds, an additional $6.08$ seconds ($4.3\%$).
This $4.3\%$ increase refers to the video-generation stage and excludes the additional costs of S1--S3.
These measurements show that the full pipeline can run end-to-end on a single H200 GPU, with a peak memory usage of about $40$~GiB.

\begin{table}[htbp]
\centering
\caption{Additional cost of dual-stream conditioning in the video-generation stage on a single NVIDIA H200 GPU. The backbone and our model use 48 and 144 input latent channels, respectively.}
\label{tab:conditioning_overhead}
\setlength{\tabcolsep}{3pt}
\begin{tabularx}{\linewidth}{@{}>{\raggedright\arraybackslash}Xrrr@{}}
\toprule
\textbf{Metric} & \textbf{Base TI2V-5B} & \textbf{Ours} & \textbf{Additional Overhead} \\
\midrule
Peak VRAM & 39.59~GiB & 39.67~GiB & $+84$~MiB \\
35 denoising steps & 133.54~s & 133.74~s & $+0.20$~s \\
Dual-stream tracking encoding & -- & 5.95~s & $+5.95$~s \\
\midrule
Total generation time & 144.20~s & 150.28~s & $+6.08$~s ($+4.3\%$) \\
\bottomrule
\end{tabularx}
\end{table}

\subsection{Input Image Lifting Construction Process.}
We first apply the monocular depth estimation model MoGe-2~\citep{wang2025moge} to the input image to predict a pixel-level 3D point map, which provides a metric geometric prior for the scene.
Based on this geometry, we use SAM3~\citep{carion2025sam3} to segment foreground objects with text prompts and obtain an object-level binary mask for each instance.
For each segmented region, SAM 3D Objects~\citep{chen2025sam3d} performs single-image 3D reconstruction conditioned on the MoGe point map.
The predicted pose directly encodes the real position and scale of the object with respect to the camera, removing the need for additional depth alignment or scale calibration.
Finally, we use the estimated poses to transform all objects from their local coordinate systems into a unified scene coordinate system.
The ground plane is estimated by applying RANSAC plane fitting to the MoGe point cloud from the lower half of the image.
We then concatenate the 3D Gaussian parameters of all objects along the object dimension to form a complete multi-object scene representation.

\subsection{Visible–Invisible Motion Field Construction.}
\label{app:vimf}
\paragraph{Training.}
We describe the construction of the visible and future-visible motion signals used in our training data construction.
\begin{itemize}[leftmargin=*,itemsep=-0.5pt]
    \item \textbf{Visible tracking video generation. }We first apply VGGT4Track~\citep{wang2025vggt} to the input RGB video to estimate per-frame depth maps, camera intrinsics and extrinsics, and depth confidence. 
    We then uniformly sample 4,900 query points on a regular grid over the first frame and track them across all subsequent frames using SpaTrack2~\citep{xiao2025spatialtrackerv2}, which outputs the 2D projected position, visibility, and confidence of each point at every frame. For rendering, only visible points within the image boundary are drawn. 
    Each point is colored by encoding its spatial position into RGB channels (R = X, G = Y, B = 1/Z), where the color is determined at the first frame and remains fixed throughout the video.

    \item \textbf{Future-visible tracking video generation. }
    For each subsequent frame $t$, we uniformly sample 8,100 candidate points on a regular grid and back-project their 3D coordinates (obtained from the VGGT4Track depth and camera pose estimates) onto the first frame via geometric reprojection. A candidate point is identified as future-visible if its projection falls outside the first-frame boundary or is occluded by the first-frame depth, indicating that it corresponds to a newly revealed region not visible in the first frame. These future-visible points are then tracked using SpaTrack2 and rendered in the same manner as described above.

\end{itemize}

\paragraph{Inference.}
We describe the construction of the visible and future-visible motion signals yielded from PGDS during inference.
\begin{itemize}[leftmargin=*,itemsep=-0.5pt]
    \item \textbf{Visible Tracking Video.}
    We first identify visible mesh vertices in the first frame via backface culling and viewport clipping, then establish a uniform grid of sampling points on the image plane and associate each point with its nearest visible vertex using a KD-tree, establishing correspondences between 2D sampling points and 3D surface points. For each subsequent frame, we solve for the rigid-body transformation via SVD using the particle positions from the rollout, transform the first-frame vertices to the current frame, and project them onto the image plane to obtain per-frame 2D trajectories. Visibility is determined by jointly evaluating viewport bounds, backface culling, and depth-map-based inter-object occlusion; only points satisfying all visibility criteria are rendered.
    \item \textbf{Future Visible Tracking Video.}
    For mesh vertices that fail the visibility check in the first frame, we scan through subsequent simulation steps frame by frame and record the first frame at which each vertex becomes visible (the query frame). To control density, we perform grid-based downsampling at each vertex's 2D projected position at its query frame, retaining only the point with the earliest query frame per grid cell. The motion tracking and visibility computation pipeline is identical to that of the visible tracking video. Additionally, for each point, visibility is set to zero in all frames preceding its query frame, and its position is locked to the query-frame value, eliminating trajectory discontinuities before appearance.
\end{itemize}
Both types of videos are rendered in the SpaTracker format (rainbow colormap assigned by first-frame Y-coordinate on a grayscale background) and the DaS format (RGB channels encoding normalized coordinates and inverse depth on a black background), with depth-sorted drawing order to handle occlusion.

\section{In-depth Analyses}

\begin{figure}[!t]
    \centering
    \includegraphics[width=0.99\linewidth]{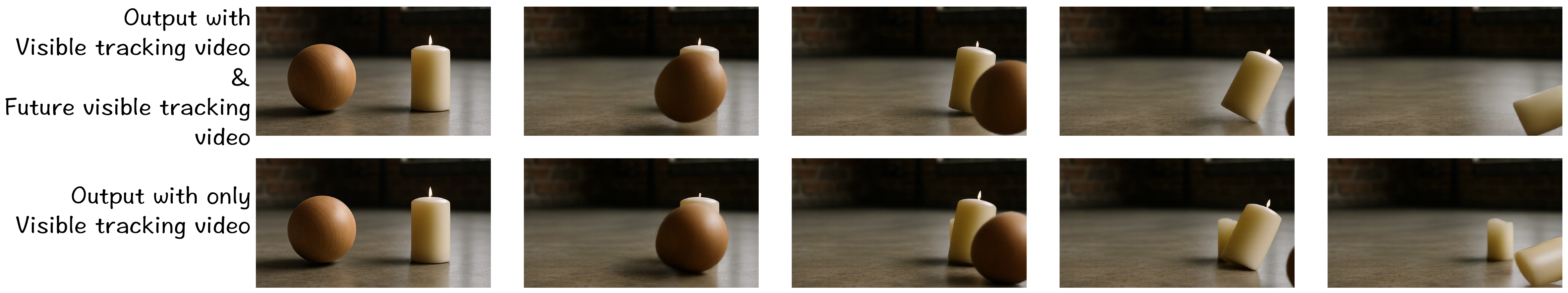}
    \vspace{-2mm}
    \caption{\textbf{Ablation on future-emergent motion signals.}
Top: our full model with both reference-visible and future-emergent tracking signals.
Bottom: model using only reference-visible tracking.}
    \label{fig:motion-ablation}
    \vspace{-2mm}
\end{figure}

\paragraph{Analysis of Unseen/Future Dynamics.}
We present an ablation study on motion conditioning signals in Figure~\ref{fig:motion-ablation}. 
The top row shows results from the full model using both reference-visible and future-emergent tracking signals, while the bottom row uses only reference-visible tracking.
When relying solely on visible tracking, the model often produces artifacts such as ghosting or missing content. 
This is because visible trajectories only describe motion of pixels observed in the reference frame, and provide no information about newly exposed regions after object movement or occlusion.
In contrast, incorporating future-emergent tracking signals provides complementary motion cues for regions that become visible over time. 
This allows the model to better reconstruct newly appearing content and maintain coherent motion dynamics.
These results highlight the necessity of modeling future-visible trajectories, as they preserve more complete motion information and lead to more stable and physically consistent video generation.

\begin{figure}[!t]
    \centering
    \includegraphics[width=0.99\linewidth]{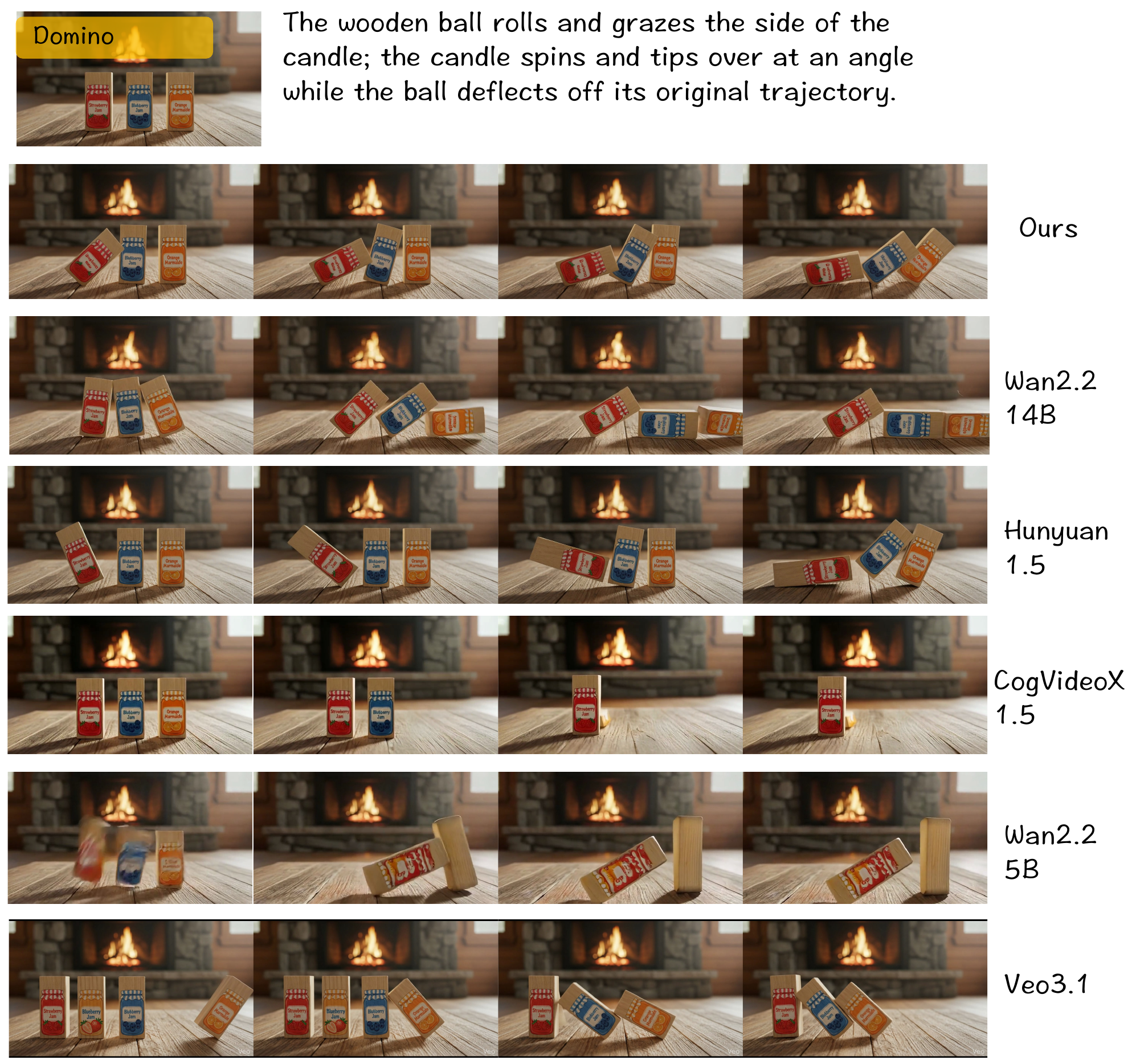}
    \vspace{-2mm}
    \caption{Qualitative comparison of domino motion.}
    \label{fig:case-1}
    \vspace{-2mm}
\end{figure}

\begin{figure}[!t]
    \centering
    \includegraphics[width=0.99\linewidth]{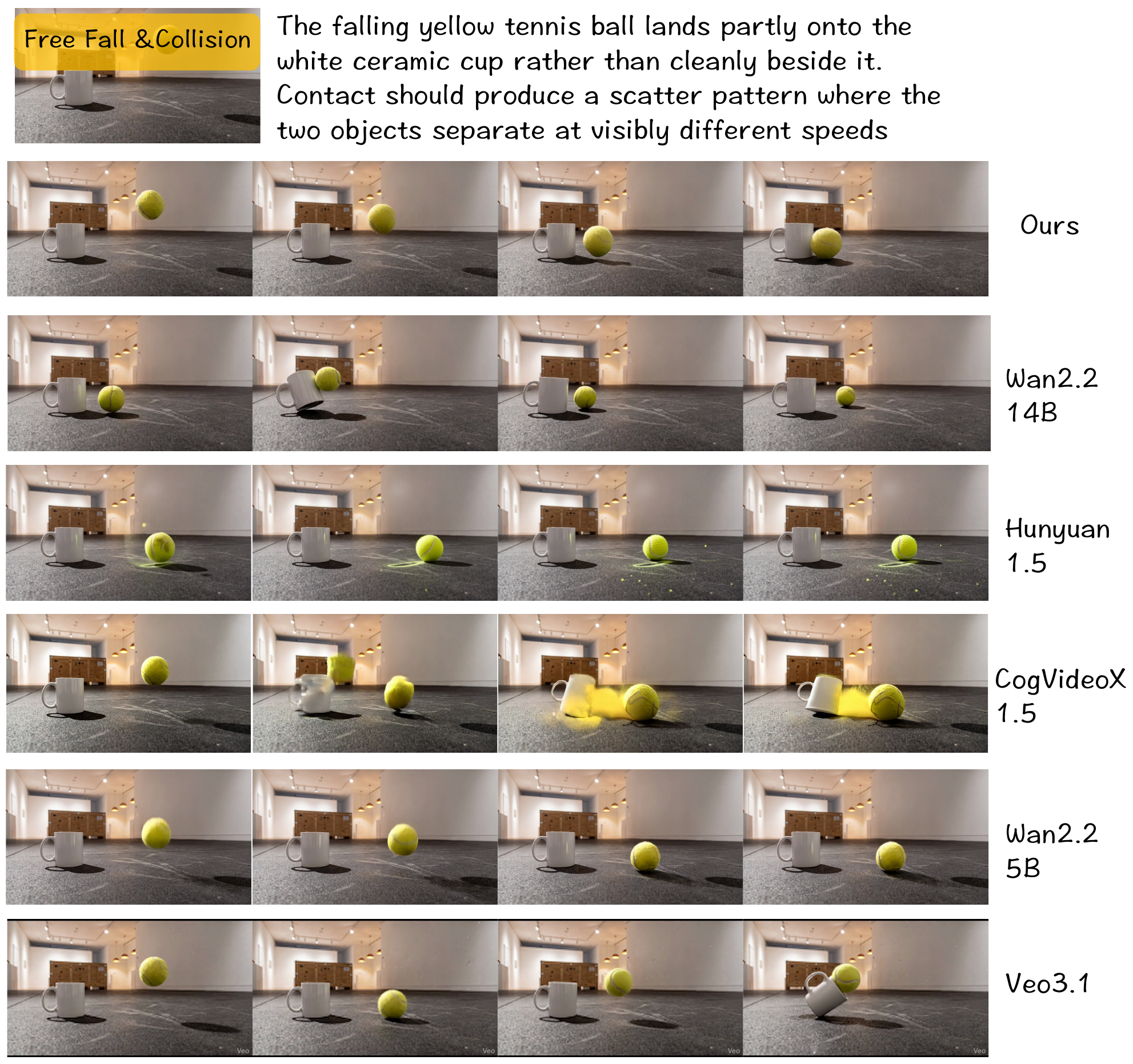}
    \vspace{-2mm}
    \caption{Qualitative comparison on free fall \& collision motion.}
    \label{fig:case-2}
    \vspace{-2mm}
\end{figure}

\begin{figure}[!t]
    \centering
    \includegraphics[width=0.99\linewidth]{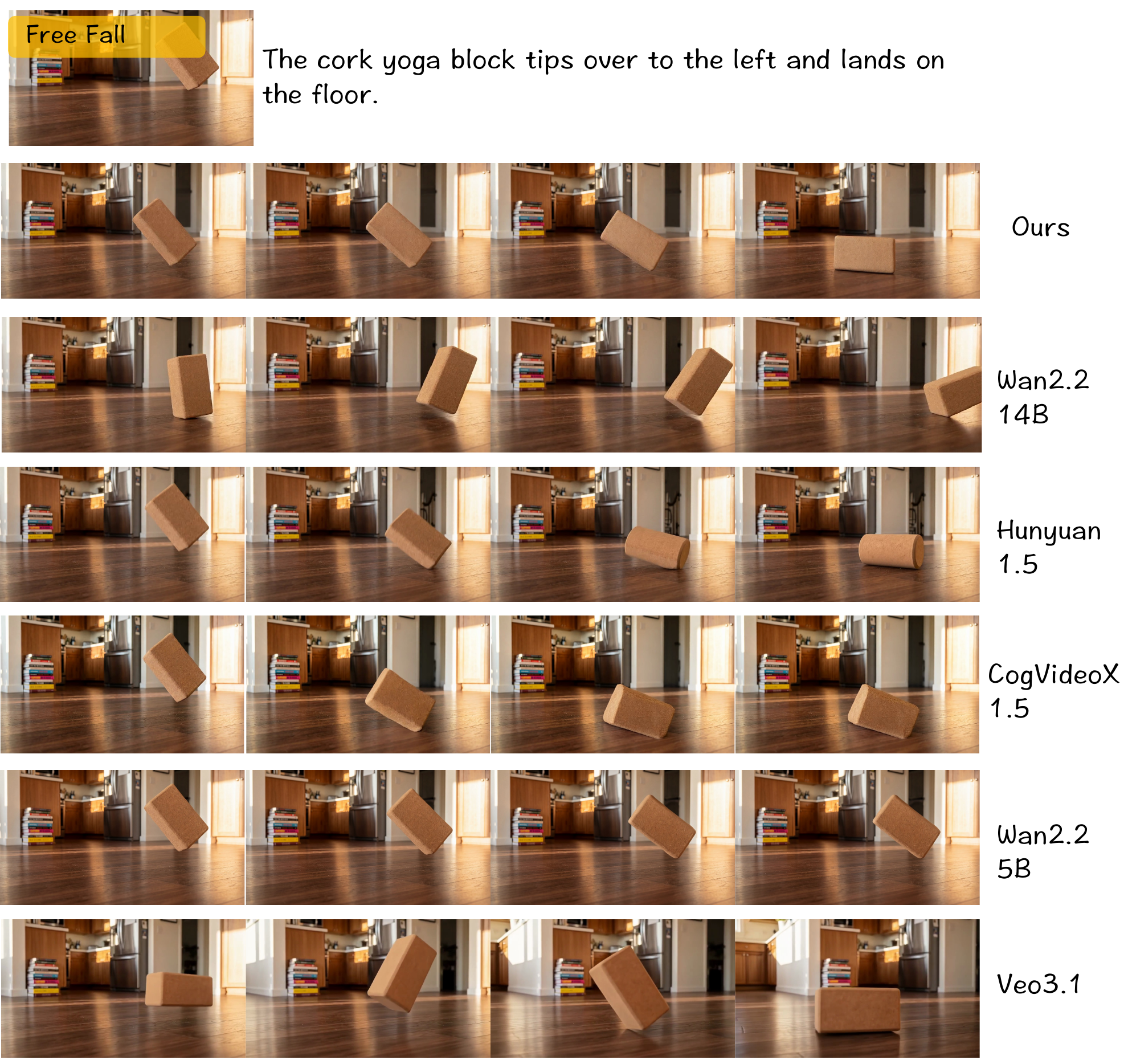}
    \vspace{-2mm}
    \caption{Qualitative comparison of free-fall motion.}
    \label{fig:case-3}
    \vspace{-2mm}
\end{figure}

\begin{figure}[!t]
    \centering
    \includegraphics[width=0.99\linewidth]{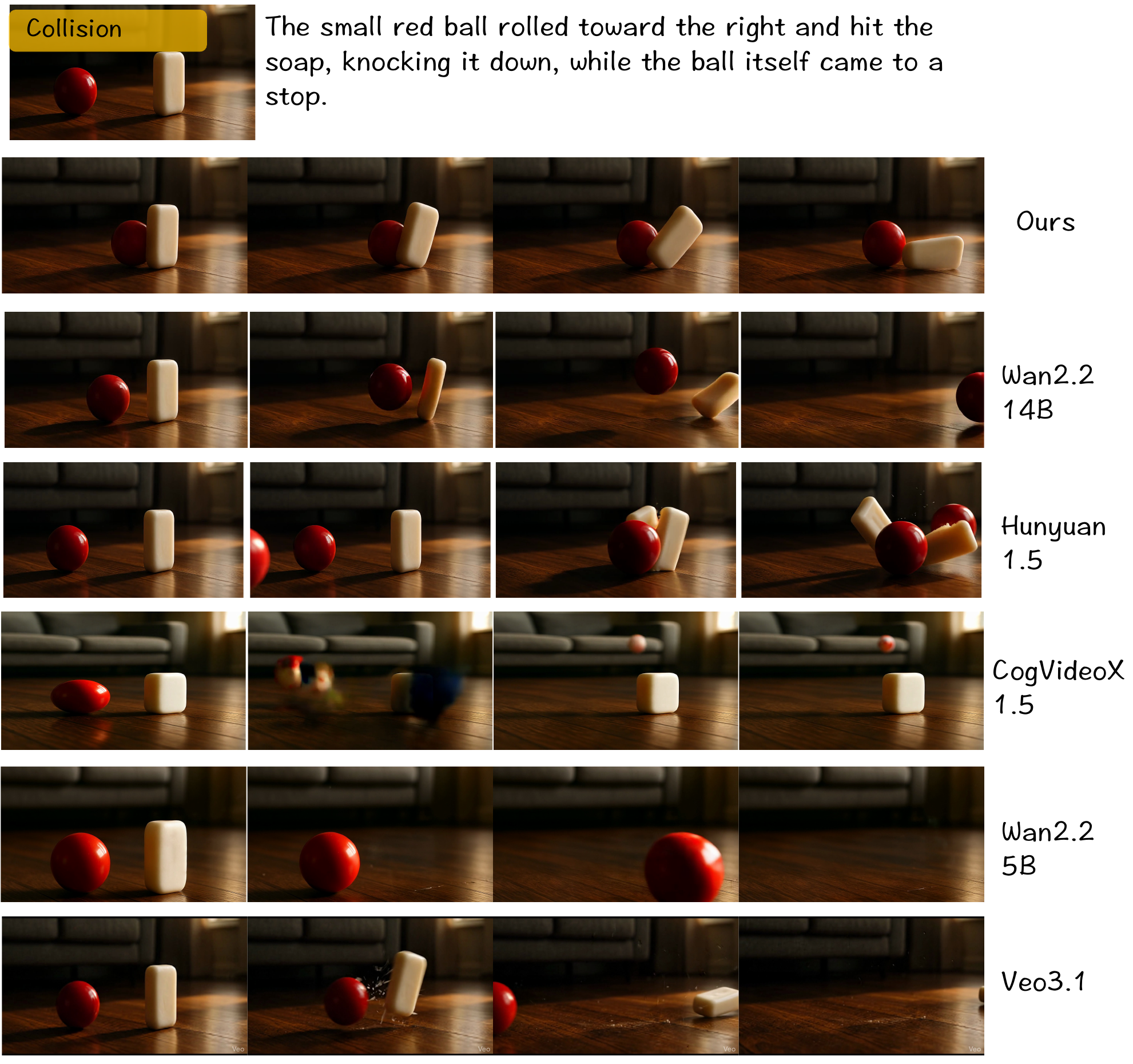}
    \vspace{-2mm}
    \caption{Qualitative comparison of collision motion.}
    \label{fig:case-4}
    \vspace{-2mm}
\end{figure}

\begin{figure}[!t]
    \centering
    \includegraphics[width=0.99\linewidth]{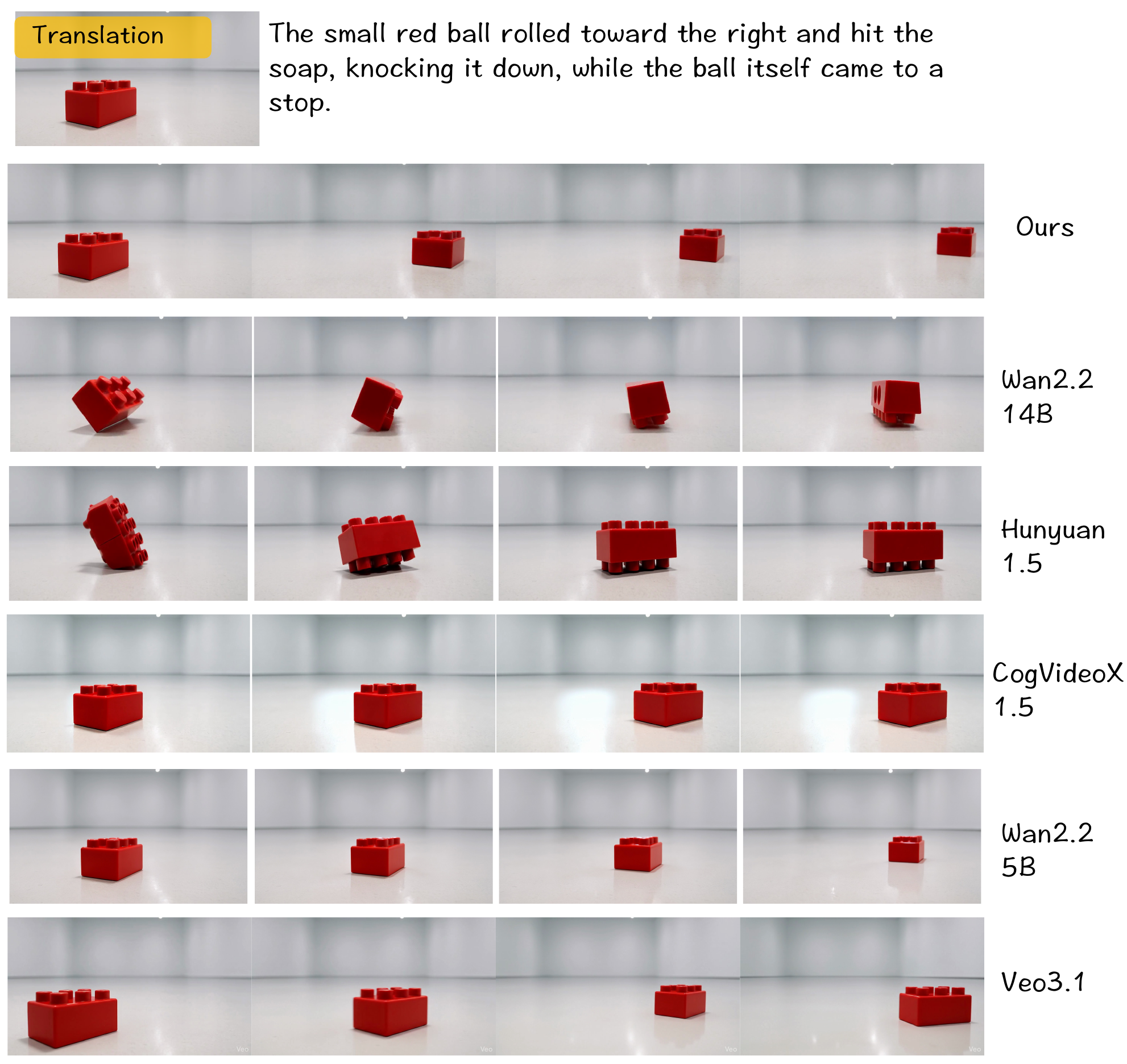}
    \vspace{-2mm}
    \caption{Qualitative comparison of translation motion.}
    \label{fig:case-5}
    \vspace{-2mm}
\end{figure}

\paragraph{More Visualizations.}
\label{app:more_vis}
We show additional visualizations compared with baselines on different scenarios in Figures~\ref{fig:case-1}-\ref{fig:case-5}.

\FloatBarrier
\section{Robustness to Upstream Perception Errors}
\label{app:perception_robustness}
We assess how incomplete motion guidance and upstream perception errors affect downstream video generation by perturbing the full pipeline in two ways: randomly dropping approximately $25\%$ of object tracks from the tracking videos, and removing $1$--$2$ objects segmented by SAM3.
We apply these perturbations separately and jointly, and compare the resulting videos with those produced by the clean full pipeline and the original Wan2.2-TI2V-5B backbone.
Table~\ref{tab:perception_robustness} reports SA and PC scores across all seven video-generation scenarios, together with their averages and relative changes from the clean model.

\begin{table}[htbp]
\centering
\caption{Robustness to incomplete motion guidance and upstream perception errors. Each entry reports SA / PC ($\uparrow$). The Tracks variant drops approximately $25\%$ of object tracks, the Objects variant removes $1$--$2$ segmented objects, and Both applies both perturbations. Coll., Dom., and F.F. denote Collision, Domino, and Free Fall. $\Delta$ reports the relative change in average SA / PC (\%) with respect to the clean full pipeline.}
\label{tab:perception_robustness}
\setlength{\tabcolsep}{2pt}
\begin{tabular*}{\linewidth}{@{\extracolsep{\fill}}lccccc@{}}
\toprule
\textbf{Scenario} & \makecell{\textbf{Ours}\\\textbf{(clean)}} & \makecell{\textbf{Ours}\\\textbf{(Tracks)}} & \makecell{\textbf{Ours}\\\textbf{(Objects)}} & \makecell{\textbf{Ours}\\\textbf{(Both)}} & \makecell{\textbf{Wan2.2}\\\textbf{TI2V-5B}} \\
\midrule
Domino & 3.976 / 4.013 & 3.777 / 3.852 & 3.578 / 3.732 & 3.459 / 3.612 & 2.650 / 3.033 \\
Collision & 4.160 / 4.394 & 3.952 / 4.218 & 3.744 / 4.086 & 3.619 / 3.955 & 2.917 / 3.067 \\
Free Fall & 4.347 / 4.328 & 4.130 / 4.155 & 4.086 / 4.155 & 3.912 / 4.025 & 2.183 / 3.083 \\
Coll.+Dom. & 3.854 / 4.016 & 3.661 / 3.855 & 3.469 / 3.735 & 3.353 / 3.614 & 2.200 / 3.033 \\
F.F.+Dom. & 3.985 / 4.017 & 3.786 / 3.856 & 3.587 / 3.736 & 3.467 / 3.615 & 2.467 / 3.033 \\
F.F.+Coll. & 4.327 / 4.589 & 4.111 / 4.405 & 3.894 / 4.268 & 3.764 / 4.130 & 2.183 / 3.083 \\
Translation & 4.013 / 4.578 & 3.812 / 4.395 & 3.772 / 4.395 & 3.612 / 4.258 & 3.000 / 4.100 \\
\midrule
Average & 4.095 / 4.276 & 3.890 / 4.105 & 3.733 / 4.015 & 3.598 / 3.887 & 2.514 / 3.205 \\
$\Delta$ (\%) & -- & $-5.0$ / $-4.0$ & $-8.8$ / $-6.1$ & $-12.1$ / $-9.1$ & $-38.6$ / $-25.0$ \\
\bottomrule
\end{tabular*}
\end{table}

Dropping tracks reduces average SA and PC by $5.0\%$ and $4.0\%$, respectively, while removing segmented objects reduces them by $8.8\%$ and $6.1\%$.
Under both perturbations, the scores decrease by $12.1\%$ and $9.1\%$, reaching $3.598$ SA and $3.887$ PC.
These scores remain higher than those of the original backbone ($2.514$ SA and $3.205$ PC), and the perturbed full pipeline outperforms the backbone in every scenario on both metrics.
The results indicate that incomplete guidance degrades generation quality, while the framework retains useful motion guidance under the tested perturbations.

SpatialTracker2~\citep{xiao2025spatialtrackerv2} is used only to construct motion-supervision data for training.
We filter out low-confidence point-cloud trajectories and retain key motion points, reducing reliance on accurate reconstruction of complete object shapes in the training data.
At inference time, PGDS uses the geometry reconstructed by the 3D lifting pipeline.
Remaining failure cases include inconsistent depth estimates for objects at similar depths, which ground-particle alignment can partially alleviate.
Segmentation is also challenging in scenes with many small objects, particularly more than ten: some objects may be missed, or multiple objects may be merged into a single segment.
Merged segments have a relatively limited impact on SAM 3D reconstruction, whereas missed objects can lead to missing particle trajectories in subsequent stages.

\end{document}